%% file: main.tex
\documentclass{article}

\usepackage[utf8]{inputenc} 
\usepackage[T1]{fontenc}    
\usepackage{hyperref}       
\usepackage{url}            
\usepackage{booktabs}       
\usepackage{amsfonts}       
\usepackage{nicefrac}       
\usepackage{microtype}      
\usepackage{lipsum}
\usepackage{fancyhdr}       
\usepackage{graphicx}       

\usepackage{amsmath}
\usepackage{amssymb}
\usepackage{mathtools}
\usepackage{color}
\usepackage{amsthm}
\usepackage{caption, subcaption}
\usepackage{natbib}
\usepackage{enumitem}
\usepackage{multirow}
\usepackage{multicol}
\usepackage{rotating}
\usepackage{array}
\usepackage{hyperref}
\usepackage{adjustbox}
\newcolumntype{C}[1]{>{\centering\arraybackslash}m{#1}}

\usepackage{arxivcustom}

\title{Stable Long-Horizon Spatiotemporal Prediction on Meshes Using Latent Multiscale Recurrent Graph Neural Networks
}

\author{
  Lionel Salesses, Larbi Arbaoui, Tariq Benamara, Arnaud Francois, Caroline Sainvitu \\
  Cenaero \\
  Gosselies, Belgium\\
  \texttt{lionel.salesses@cenaero.be} \\
}

\definecolor{mydarkblue}{rgb}{0,0.08,0.45}
\hypersetup{ %
    pdftitle={Stable Long-Horizon Spatiotemporal Prediction on Meshes Using Latent Multiscale Recurrent Graph Neural Networks},
    pdfauthor={Lionel Salesses},
    pdfkeywords={Long-Horizon Prediction, Multiscale Modeling, Additive Manufacturing, Scientific Machine Learning, Graph Autoencoder},
    colorlinks=true,
    linkcolor=mydarkblue,
    citecolor=mydarkblue,
    filecolor=mydarkblue,
    urlcolor=mydarkblue,
    }

\makeatletter
\def\placekeywords{\xdef\@thefnmark{}\@footnotetext}
\makeatother

\begin{document}

\flushbottom \twocolumn[
  \maketitle
]
\sloppy

\placekeywords{\keywords{Long-Horizon Prediction \and Multiscale Modeling \and Additive Manufacturing \and Scientific Machine Learning \and Graph Autoencoder}}

\input{abstract}
\input{Intro}
\input{RelatedWork}
\input{Dataset}

\input{Methodology}
\input{Results}
\input{Conclusion}

\section*{Impact Statement}
This paper presents work whose goal is to advance the field of Machine
Learning. There are many potential societal consequences of our work, none
which we feel must be specifically highlighted here.

\section*{Acknowledgments}
This work is supported by the “ARIAC by DigitalWallonia4.ai” reasearch project (grant agreement No 2010235 – TRAIL institute) and benefited from computational resources made available on Lucia, the Tier-1 supercomputer of the Walloon Region, infrastructure funded by the Walloon Region (grant agreement No 1910247). This work is supported by the European Regional Development Fund (ERDF/FEDER) and the Walloon Region of Belgium through project 364 Cenaero\_AdviseAM (programme 2021-2027).

\bibliographystyle{unsrt}  
\bibliography{paper}  

\newpage
\appendix
\onecolumn

\input{Appendix}

\end{document}

%% file: abstract.tex
\begin{abstract}
Accurate long-horizon prediction of spatiotemporal fields on complex geometries is a fundamental challenge in scientific machine learning, with applications such as additive manufacturing where temperature histories govern defect formation and mechanical properties. High-fidelity simulations are accurate but computationally costly, and despite recent advances, machine learning methods remain challenged by long-horizon temperature and gradient prediction.
We propose a deep learning framework for predicting full temperature histories directly on meshes, conditioned on geometry and process parameters, while maintaining stability over thousands of time steps and generalizing across heterogeneous geometries.
The framework adopts a temporal multiscale architecture composed of two coupled models operating at complementary time scales. Both models rely on a latent recurrent graph neural network to capture spatiotemporal dynamics on meshes, while a variational graph autoencoder provides a compact latent representation that reduces memory usage and improves training stability.
Experiments on simulated powder bed fusion data demonstrate accurate and temporally stable long-horizon predictions across diverse geometries, outperforming existing baseline. Although evaluated in two dimensions, the framework is general and extensible to physics-driven systems with multiscale dynamics and to three-dimensional geometries.
\end{abstract}

%% file: Intro.tex
\section{Introduction}
Accurate long-horizon prediction of spatiotemporal fields defined on complex and evolving geometries is a central challenge in scientific machine learning. Such problems arise across a wide range of physical systems, including additive manufacturing \cite{tian2025physics}, climate and weather modeling \cite{verma2024climode}, \cite{price2025probabilistic}, and fluid dynamics \cite{gao2024towards}, where system dynamics exhibit strong coupling between fast local phenomena and slow global evolution. Learning models that are both temporally stable over long horizons and computationally efficient remains an open problem, particularly when predictions must be made directly on irregular meshes \cite{zhao2019t}, \cite{yue2024tgn}.
Additive manufacturing provides a representative and practically relevant instance of this challenge. During metal powder bed fusion, the temperature history governs defect formation, residual stresses, and final mechanical properties (see \cite{narasimharaju2022comprehensive} and \cite{bian2025review}). While high-fidelity numerical solvers can accurately capture these thermal dynamics, their computational cost precludes real-time inference and limits large-scale design exploration (see \cite{chiumenti2017numerical} and \cite{burkhardt2022thermo}). While recent machine learning approaches have shown promise in accelerating thermal prediction, achieving reliable long-range forecasts of temperature fields and gradients on variable geometries remains difficult in practice \cite{choi2025transfer}.
A key difficulty stems from the intrinsic temporal multiscale structure of the underlying physics. In additive manufacturing, short-term intralayer dynamics are dominated by highly localized laser-material interactions, whereas long-term interlayer dynamics are governed by heat diffusion and cumulative energy deposition. Treating these disparate time scales within a single monolithic sequence model often leads to unstable training, excessive memory usage, or error accumulation over long horizons \cite{lippe2023pde}.
In this work, we propose a Latent Multiscale Recurrent Graph Neural Network (LM-RGNN) framework that explicitly leverages this temporal separation. Our approach decomposes long-horizon prediction into two coupled but independently trained models operating at complementary time scales. Both models perform sequence prediction directly on meshes using a recurrent graph neural network, while a Variational Graph Autoencoder (VGAE) provides a compact latent representation of temperature fields. This latent formulation substantially reduces memory requirements and facilitates stable information propagation over long temporal horizons.
From a machine learning perspective, our contribution is a general latent multiscale modeling strategy for long-horizon spatiotemporal prediction on graphs, rather than a task-specific architecture. The proposed framework is designed to accommodate variable meshes, long sequences, and multiscale temporal dynamics, which are common across many physics-driven learning problems. Additive manufacturing serves as a challenging testbed that highlights these characteristics but does not constrain the applicability of the method.
We evaluate the proposed approach on simulated powder bed fusion data and demonstrate improved predictive accuracy, temporal stability, and computational efficiency compared to an existing graph-based baseline \cite{choi2025transfer}. Beyond temperature fields, the model accurately captures derived quantities such as spatial and temporal gradients and melt-pool localization. While our experiments focus on two-dimensional simulations, the framework naturally extends to three-dimensional geometries and other multiscale physical systems.
Overall, this work advances the state of long-horizon spatiotemporal modeling on graphs by introducing a latent multiscale formulation that improves stability and scalability, offering a general tool for scientific machine learning applications characterized by coupled fast and slow dynamics. Potential extensions and application domains are discussed in Appendix~\ref{appendix:general-appli}.

%% file: RelatedWork.tex
\section{Related Work}
\paragraph{Machine Learning for Spatiotemporal and Physics-Based Modeling.}
Learning long-horizon spatiotemporal dynamics governed by PDEs remains a central challenge in scientific machine learning. Autoregressive models, while widely used due to their simplicity, often suffer from error accumulation and instability over long rollouts, particularly when applied to high-dimensional physical fields. Lippe et al. \cite{lippe2023pde} show that standard autoregressive models often fail to produce stable and accurate rollout predictions due to their inability to capture the high-frequency components of PDE solutions. To address this limitation, they introduce PDE-Refiner, a diffusion-inspired refinement framework that enhances high-frequency modeling for autoregressive predictors operating on fixed regular grids through a multistep refinement strategy. The method is validated on fluid dynamics benchmarks, yielding substantial improvements in rollout accuracy over horizons of up to 1000 time steps.
Graph-based models have emerged as a natural paradigm for learning physics on unstructured domains. The EAGLE framework \cite{janny2023eagle} introduces an autoregressive graph encoder-decoder architecture combining mesh clustering with a core attention mechanism to model turbulent flow dynamics directly on simulation meshes. Trained on CFD data, EAGLE demonstrates the ability to predict flows across varying two-dimensional geometries, with evaluation over horizons of up to 250 time steps showing a gradual, though limited, error growth. Similarly, the Temporal Graph Network (TGN) framework \cite{yue2024tgn} explores recurrent graph-based architectures for physics prediction on unstructured meshes, highlighting the difficulty of achieving long-horizon stability due to the joint spatialtemporal complexity of physical systems.
Recent work has also explored attention-based and operator-learning approaches for spatiotemporal modeling. The ASNO framework \cite{karkaria2025asno} proposes an autoregressive architecture that decouples spatial and temporal modeling using attention mechanisms, followed by a neural operator to predict spatiotemporal fields. ASNO is evaluated on benchmark PDE systems, including melt pool prediction in additive manufacturing. Operating on fixed grids, it reports stable rollouts up to 95 time steps with reduced error accumulation compared to baselines. 
\paragraph{Machine Learning for Additive Manufacturing and Thermal Modeling.}
Several studies in additive manufacturing employ Physics-Informed Neural Networks (PINNs) to achieve accurate thermal predictions for a specific geometries \cite{zhu2021machine,liao2023hybrid}, achieving good melt pool reconstruction but typically focusing on limited geometries and short prediction horizons.
Tian et al. \cite{tian2025physics} combine physics-informed RNNs with convolutional LSTMs to autoregressively predict thermal histories on regular grids for thin-wall geometries, achieving an RMSE of about 20 °C over short-term horizons (11.25 s), demonstrating the potential of recurrent formulations for thermal modeling, but without evaluation in long-horizon rollout settings.
Graph-based learning has recently gained traction in additive manufacturing due to its ability to operate directly on variable part meshes. Choi et al. \cite{choi2025transfer} propose a GNN-RNN architecture trained on additive manufacturing simulations to predict thermal histories for unseen complex three-dimensional geometries. Operating at the subsequence level, their model achieves accurate and stable predictions over horizons of up to 50 time steps. Considering to its strong performance and close methodological relevance, this approach is adopted as a baseline in the present work and detailed in Appendix \ref{appendix:dec-rgnn}.
\paragraph{Positioning of This Work.}
In contrast to prior approaches, our work targets long-horizon spatiotemporal prediction on irregular meshes by explicitly leveraging the temporal multiscale structure of the underlying physics. Unlike single-scale autoregressive graph models such as EAGLE \cite{janny2023eagle} or subsequence-based approaches in additive manufacturing \cite{choi2025transfer}, we decompose the prediction task into coupled interlayer and intralayer models operating at complementary time scales. This design enables stable prediction over thousands of time steps while remaining computationally efficient.
Moreover, whereas diffusion-based refinement strategies \cite{lippe2023pde} and neural operator approaches \cite{karkaria2025asno} primarily operate on fixed grids, our framework directly models fields on variable meshes using a latent recurrent graph neural network. 
While evaluated in the context of additive manufacturing, the proposed latent multiscale RGNN framework is general and applicable to a broad class of physics-driven systems characterized by coupled fast local dynamics and slow global evolution.

%% file: Dataset.tex
\section{Additive Manufacturing Benchmark and Dataset}
\begin{figure}[!htb]
    \centering
    \includegraphics[width=\linewidth]{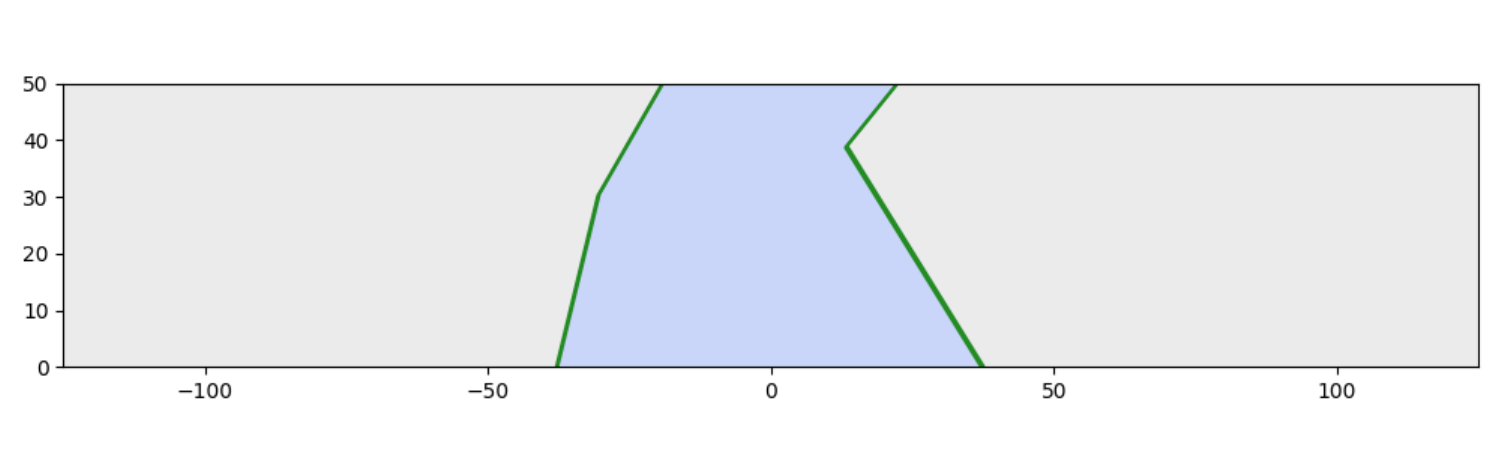}
    \caption{Example of printed geometry. Metallic powder is in gray, the final printed part is in blue and the part boundaries are highlighted in green.}
    \label{fig:geometry}
\end{figure}

The task addressed in this work is to predict the complete temperature history during the additive manufacturing process using a deep learning approach capable of handling variable part geometry. The present study focused on two-dimensional geometries, which already pose significant modeling challenges; however, the proposed machine learning framework is explicitly designed with scalability to three-dimensional parts in mind, where data generation costs, memory requirements, and computational complexities become substantially higher.
The task can be formulated around the following objectives: 
\begin{enumerate}[label=(\roman*)]
    \item Achieve temporally stable long-horizon predictions of the full temperature evolution defined on geometry-specific simulation meshes. \label{obj:1}
    \item Accurately predict temperature fields with respect to physically and application-relevant metrics, including spatial and temporal temperature gradients that are critical to printed part quality (see \cite{ali2018residual} or \cite{abolhasani2019analysis}). \label{obj:2}
    \item Generalize effectively to geometries unseen during training.  \label{obj:3}
    \item Control memory consumption to ensure scalability and to anticipate the significant increase in memory footprint of three-dimensional simulations. \label{obj:4}
    \item Ensure inference times is compatible with near real-time prediction, enabling future applications such as process monitoring. \label{obj:5}
\end{enumerate}
The dataset is derived from two-dimensional numerical simulations performed using a finite element-based thermal solver. Those simulations are designed to predict spatial and temporal evolution of the temperature during additive manufacturing of metallic parts with a particular focus on the powder bed fusion process. 
Numerical simulations are based on a layer by layer progressive approach that employs mesh element activation techniques. For each layer, the thermal evolution is characterized by a distinct heating and cooling phase. During the heating phase, thermal responses are computed using a moving Gaussian heat source with an alternating linear laser path. Each activated layer is composed of a metallic powder region and a solid metal region targeted by the laser heat source. More details on the manufacturing process and the modeling hypothesis can be found in Appendix \ref{appendix:dataset-details}.
The dataset is designed to train and develop a deep learning approach focusing on geometrical characteristics. Process parameters including laser course speed, laser power, cooling time and scan trajectory are kept fixed throughout this study. 
Each simulation is associated with a parameterized geometry that satisfies additive manufacturability constraints, as detailed in Appendix~\ref{appendix:subsec:geo-param}. An example of such a geometry is shown in Figure~\ref{fig:geometry}. Additional details on the dataset generation and the underlying simulation model are provided in Appendix~\ref{appendix:subsec:dataset-desc} and Appendix~\ref{appendix:subsec:sim-model}, respectively.

\begin{figure}[!htb]
    \centering
    \begin{subfigure}[b]{\linewidth}
        \centering
        \includegraphics[width=0.98\linewidth]{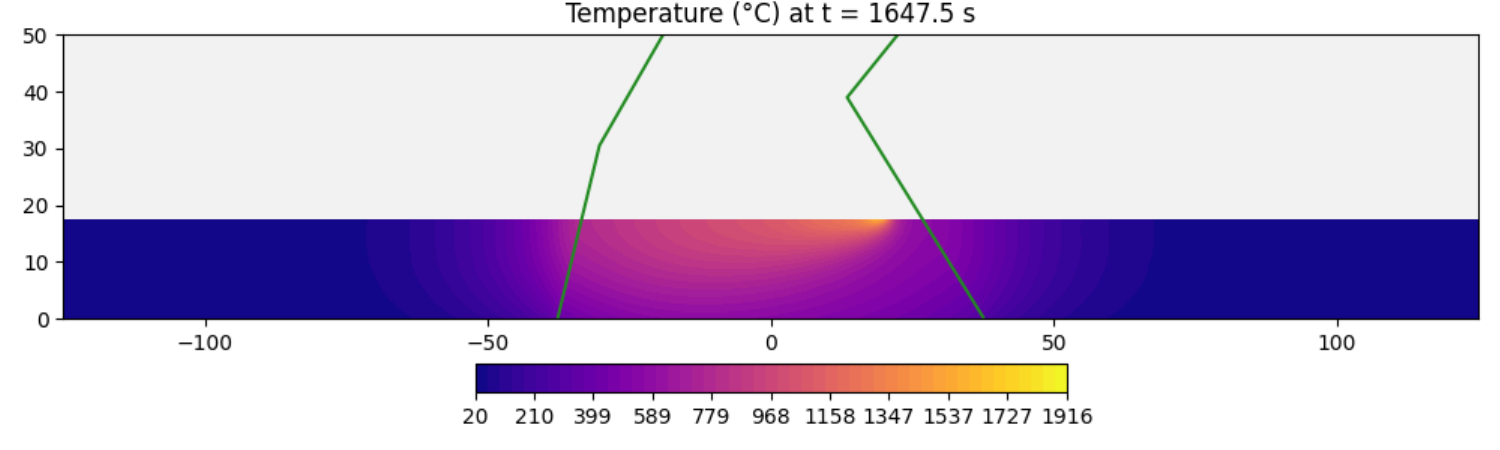}
        \label{fig:temperature_example:a}
    \end{subfigure}\hfill
    \begin{subfigure}[b]{\linewidth}
        \centering
        \includegraphics[width=0.98\linewidth]{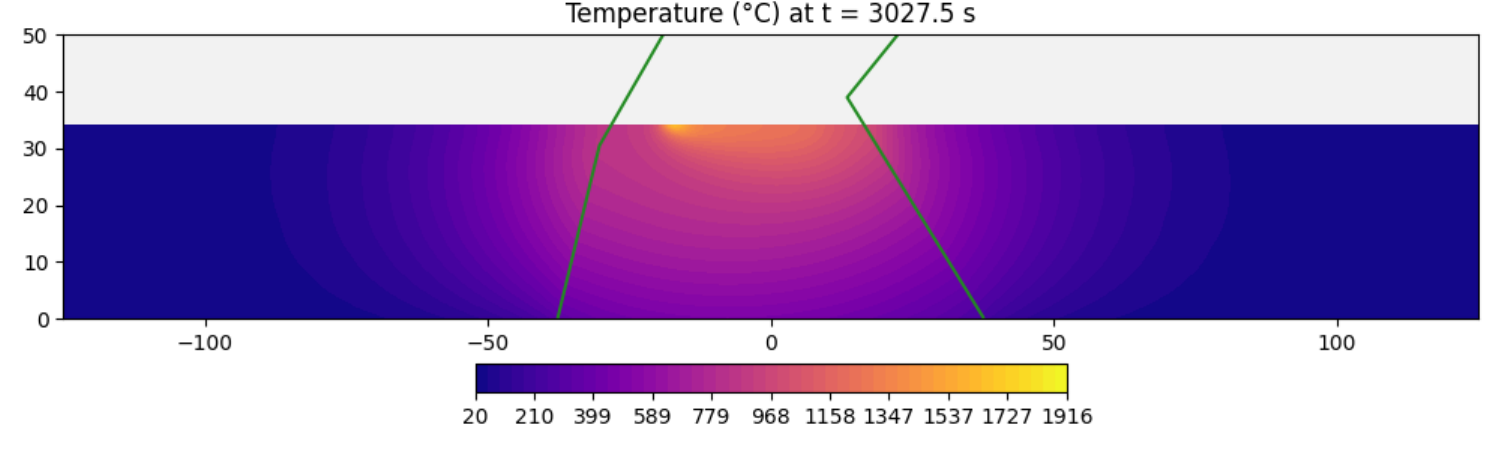}
        \label{fig:temperature_example:b}
    \end{subfigure}\hfill
    \begin{subfigure}[b]{\linewidth}
        \centering
        \includegraphics[width=0.98\linewidth]{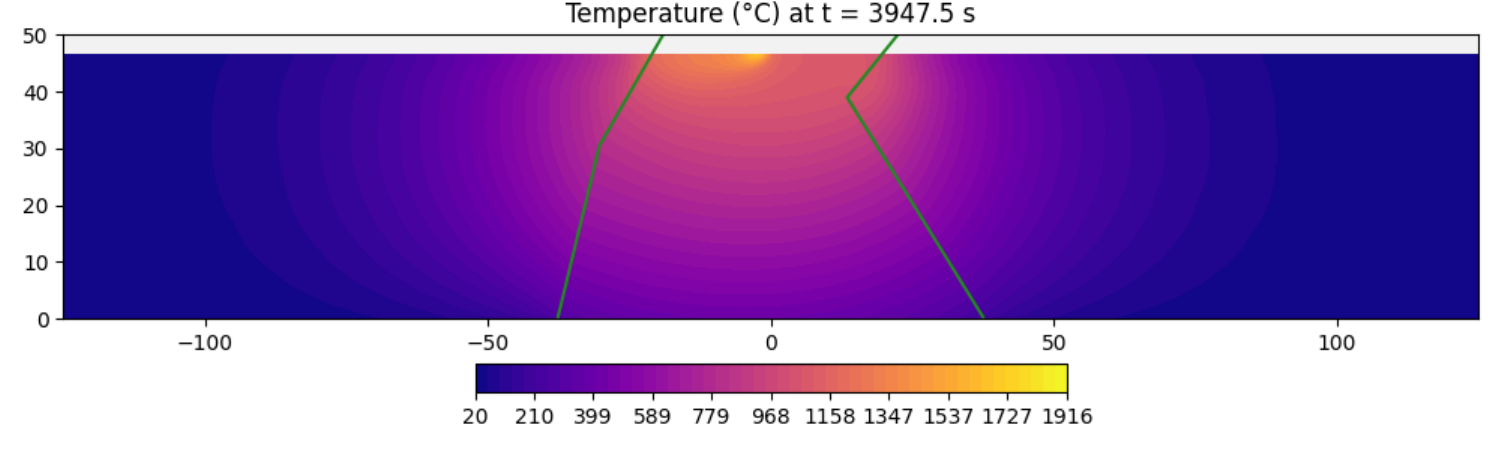}
        \label{fig:temperature_example:c}
    \end{subfigure}
    \caption{Temperature field snapshots for a fixed printed geometry at multiple time steps. The green lines indicate the part boundaries, enclosing the solid metallic region. The temperature is simulated for both metallic part and the surrounding powder.}
    \label{fig:temperature_example}
\end{figure}

For each simulation in the dataset, we have access to the computational mesh, including its topology, node coordinates, and material mask, as well as the laser process parameters, namely the time-resolved laser position and power. The simulated temperature field is provided at each simulation time step for all currently unmasked nodes. Since the simulation domain increases progressively in a layer-by-layer manner during the printing process, an active-node mask is available at every time step. 
Temperature fields are defined at mesh nodes and recorded for each simulation time step (see Figure \ref{fig:temperature_example} for representative snapshots). The length of the resulting temperature sequences depends on the part geometry and, in the present dataset, ranges from 6600 to 12000 time steps (corresponding to approximately 1600s to 3000s). Temperature values span from a minimum of 20°C to a maximum of approximately 2100°C. The highest temperatures occur in the vicinity of the laser beam, where metallic powder melts down to form the melt pool. This region is characterized by steep temperature gradients and complex physical processes, including phase change, which are modeled through the inclusion of a latent heat term in the governing equations.
The 140 generated simulations are randomly partitioned into 80 simulations for training, 20 for validation, and 40 for testing.

%% file: Methodology.tex
\section{Latent Multiscale Recurrent Graph Architecture}
In this section, we introduce an architecture for predicting the temperature history of a printed part from process parameters and geometric information. The proposed approach adopts a time-multiscale modeling strategy based on two coupled models, each operating at a distinct temporal scale. Both models are built upon two core components: a RGNN, which enables the sequential modeling of temperature fields defined on a mesh, and a VGAE, which provides a compact latent representation of the temperature fields.

\subsection{Architecture Design Motivations}
Since the temperature fields are defined on simulation meshes, it is natural and advantageous to operate directly on mesh-based data using GNNs, thereby avoiding interpolation onto a fixed snapshot grid, which would introduce additional approximation errors. In particular, mesh representations more accurately capture part interfaces, whereas image-based interpolations tend to blur these interfaces, leading to aliasing and loss of geometric fidelity. GNNs enable the processing of both local mesh connectivity and global structural features, allowing them to simultaneously model fine-scale details and the overall geometry of the part. As a result, GNNs are well suited to this class of problems.
The objective of this work is to learn a mapping from the process parameters (specifically the laser path) to time-resolved sequences of temperature fields over very long temporal horizons. This requirement necessitates a model with strong temporal stability and minimal error accumulation. As discussed in Appendix \ref{appendix:sec:RNN-motivations}, RNN are well suited to this class of problems; however, this approach poses significant computational challenges. In particular, the memory footprint required for backpropagation through long sequences becomes prohibitive. To address this limitation, we adopt Truncated Backpropagation Through Time (TBPTT) \cite{jaeger2002tutorial}, which restricts backpropagation to a limited number of time steps while allowing the forward pass to proceed sequentially. This strategy substantially reduces memory usage while maintaining the ability to model long-term temporal dependencies. A second major challenge arises from the vanishing and exploding gradient phenomena commonly encountered in recurrent neural networks. To mitigate these effects, we employ an interlayer and intralayer splitting strategy (detailed below), which allows the RNN to be trained on coarser temporal sequences. 
This strategy enables the interlayer and intralayer models to be trained independently, facilitating modular development and more effective prototyping. 
To extend recurrent neural networks to graph-structured data, we introduce a RGNN architecture that combines a standard recurrent unit (specifically a Gated Recurrent Unit (GRU)) with graph neural network layers. For computational efficiency, the RGNN operates on a low-dimensional latent representation of the temperature fields rather than the full high-dimensional state
, substantially reducing memory usage and alleviating GPU constraints on batch size and TBPTT sequence length. The temperature fields defined on complex geometries exhibit multiple sources of variability, including global geometric structure and highly localized thermal dynamics associated with the laser-induced melt pool. To obtain such latent representations, we employ a VGAE, which generalizes the classical VAE to graph-structured data (see \cite{kipf2016variational} and \cite{valencia2025learning}). The use of hierarchical graph pooling and unpooling operations allows the VGAE to extract multiscale features from both the part geometry and the temperature field, making it well suited for downstream recurrent modeling.
We now provide a detailed description of the components that constitute the proposed architecture.

\subsection{GraphGRU: Graph Gated Recurrent Unit}
The standard formulation of the GRU operates on vector-valued inputs and hidden states. This formulation has been extended to structured data such as images and graphs by replacing linear transformations with convolutional or graph-based operators \cite{seo2018structured}. In this work, we generalize the GRU to graph-structured data by replacing the matrix multiplications in the classical GRU with Message-Passing Graph Neural Networks (MPGNNs).
Implementation details of the resulting GraphGRU module are provided in Appendix~\ref{appendix:subsec:graphGRU}.
\subsection{RGNN: Recurrent Graph Neural Network} \label{sec:RGNN}
Let $T_0$ denote the initial temperature and $M$ the part mesh, which is represented as a graph $(A, E)$, where $A$ is the adjacency matrix and $E$ contains edge features encoding geometric information such as edge lengths and unit direction vectors. The initial hidden state $H_0$ is constructed by normalizing the initial temperature field and padding it with zeros to match the recurrent hidden state size. 
At each time step $t$, the model is provided with an active-node mask $m_t$, a material mask $\mu_t$ indicating the local material state (metal or powder), and a vector of global attributes $g_{t+1}$ that encodes process parameters, including changes in laser position and power since the previous time step.
The RGNN evolves according to the following recurrent update (see Figure \ref{fig:rgnn}):
\begin{equation}
    \begin{aligned}
        &X_{t+1} = m_{t+1} \oplus \mu_{t+1}, \\ 
        &H_{t+1} = \mathrm{GraphGRU}(X_{t+1}, H_t, g_{t+1}) \odot m_{t+1}, \\
        &T_{t+1} = H_{t+1}[0],
    \end{aligned}
\end{equation}
where the hidden state $H_t$ is a node-feature tensor whose first component corresponds to the nodal temperature $T_t$.

\subsection{VGAE: Variational Graph Auto-Encoder} \label{sec:VGAE}
The VGAE takes as input a temperature field $T_{true}$ defined on a mesh, represented as a graph $(A, E)$, together with a mask of active nodes $m$. Its purpose is to encode the temperature field into a compact latent representation defined on a reduced, latent graph. 
The learned latent representation enables downstream models to operate directly in latent space, substantially reducing memory consumption. 
Implementation details of the VGAE module are provided in Appendix~\ref{appendix:sec:VGAE}.

\subsection{Latent-RGNN}
The RGNN introduced in Section \ref{sec:RGNN} can be extended to operate directly in latent space. The recurrent hidden state may be initialized either to zero or by encoding the initial temperature field using a pre-trained VGAE encoder. In the latter case, the initial latent state is set to the node-wise mean of the latent distribution predicted by the encoder. The latent-RGNN produces a sequence of latent representations, which are subsequently decoded into temperature fields using the pre-trained VGAE decoder with fixed parameters. 
The primary motivation for the latent-RGNN design is to reduce GPU memory consumption and avoid computationally expensive operations on high-dimensional fields. More details are given is Appendix \ref{appendix:sec:latent-RGNN-arch}. 

\subsection{Time-Multiscale Strategy} 
\begin{figure*}[!htb]
    \centering
    \includegraphics[width=0.97\textwidth]{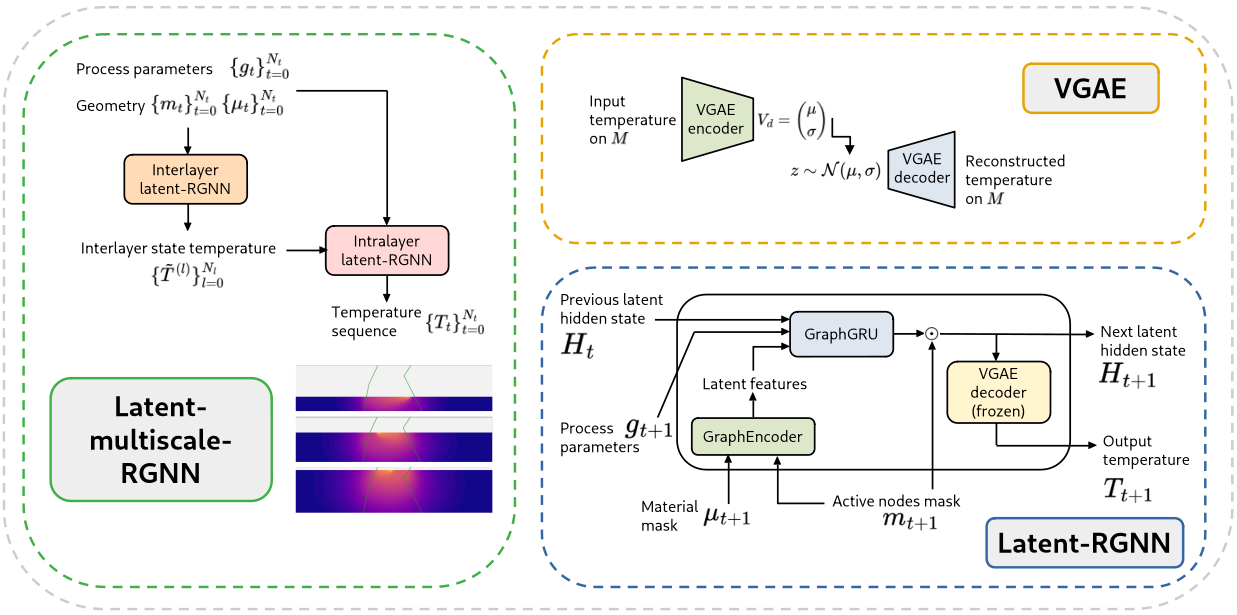}
    \caption{Overview of the LM-RGNN architecture and its main components, including the latent-RGNN and the VGAE.}
    \label{fig:framework}
\end{figure*}
The layer-by-layer nature of the additive manufacturing process naturally induces two distinct temporal scales: (i) an \emph{interlayer} scale, corresponding to the temperature field at the completion of each printed layer, and (ii) an \emph{intralayer} scale, corresponding to the fine-grained temporal evolution of the temperature field during the printing of a single layer. We exploit this structure by introducing a time-multiscale modeling strategy based on two coupled but independently trained models, as summarized in Figure~\ref{fig:framework}.
The interlayer model is trained to predict the temperature field at the end of each layer and capture long-term thermal dynamics dominated by heat diffusion and cumulative heat accumulation effects. Conditioned on these interlayer predictions, the intralayer model is trained to reconstruct the temperature evolution within each layer. Specifically, it takes as input the predicted initial temperature at the start of a layer, along with the layer associated process parameters, and outputs the transient temperature fields until layer completion. The intralayer model focuses on short-term dynamics associated with laser motion and melt-pool phenomena, which are characterized by large but spatially localized temperature gradients.
Both models are trained on substantially shorter temporal sequences than a full-sequence RGNN, resulting in improved training stability and reduced probability of vanishing and exploding gradient effects. In addition, this two-stage design enables more efficient inference: once the interlayer predictions are obtained, intralayer temperature evolution for all layers can be computed in parallel. This contrasts with full-sequence RGNNs, which require strictly sequential prediction across all time steps and thus incur significantly higher inference costs.
Let $\lbrace T_t \rbrace_{t=0}^N$ denote the sequence of temperature fields over time. This sequence can be partitioned according to the layer structure of the process into a collection of layer-wise subsequences $\lbrace T^{(l)} \rbrace_{l=0}^L$, where $L$ denotes the total number of layers. Each subsequence $T^{(l)} = \lbrace T_t \rbrace_{t=0}^{N_l}$, corresponds to the temperature evolution during the printing of layer $l$, with $N_l$ denoting the number of time steps associated with that layer. We further denote $\tilde{T}^{(l)}$ the temperature field after completion of printing layer $l$.
\paragraph{Interlayer Model.} The interlayer model, implemented using the latent-RGNN architecture, is trained to predict the sequence of layer-wise terminal temperature fields $\lbrace \tilde{T}^{(l)} \rbrace_{l=0}^{N_l}$. It takes as input global geometric information, including active-node masks and material masks, as well as a sequence of layer-level process descriptors such as layer width, printing duration, and cooling duration.
\paragraph{Intralayer Model.} The intralayer model, also based on the latent-RGNN architecture, is trained to predict the within-layer temperature evolution $T^{(l)}$ for each layer $l$. It is conditioned on the initial temperature field at the start of the layer, as predicted by the interlayer model, together with the same global geometric information and a sequence of time-resolved process parameters.
\paragraph{Latent-Multiscale-RGNN (LM-RGNN).}
The LM-RGNN refers to the coupled architecture obtained by integrating the interlayer and intralayer models described above (see Figure \ref{fig:framework}).

%% file: Results.tex
\section{Experiments and Results}
All details regarding the optimizer, learning rate scheduler, temperature sequence subsampling strategy, and training hyperparameters are provided in Appendix \ref{appendix:train-details}.
\subsection{Evaluation Metrics}
We assess model performance using a comprehensive set of metrics, including Mean Absolute Error (MAE), Mean Absolute Percentage Error (MAPE), Mean Maximum Error (MME), melt-pool Intersection over Union (mpIoU), and temporal MAE (t-MAE). In addition, we evaluate errors on spatial and temporal temperature gradients through the MAE of spatial gradients ($\text{MAE}_{\nabla_{xy}}$) and temporal gradients ($\text{MAE}_{\nabla_t}$). The t-MAE metric is used to quantify temporal stability of the predictions.
In addition to predictive accuracy, we report the number of model parameters, memory consumption at training-time, and per-time-step inference time measured in milliseconds. Further details on the evaluation metrics and experimental protocol are provided in Appendix~\ref{appendix:eval-metrics}. 

\subsection{Baseline}
As a baseline for comparison, we reimplemented the Decoupled-RGNN architecture proposed by \cite{choi2025transfer}. This model is designed to predict the temperature evolution throughout the printing process and to generalize across previously unseen geometries. Additional details regarding the architecture and our implementation are provided in Appendix \ref{appendix:dec-rgnn}.

\subsection{VGAE Model Performance} \label{sec:perf-gvae}
The VGAE model performances are reported in Appendix \ref{appendix:sec:VGAE-perfs}.
For all latent-based models considered in the following, we employ a VGAE trained with a latent dimensionality of 16, which produced the best predictive accuracy.
Overall, the VGAE exhibits favorable performance with respect to the considered objectives, yielding low reconstruction error, stable behavior over time, satisfactory accurate predictions of melt pool extent, and controlled worst-case errors.

\subsection{LM-RGNN Model Performance}

\begin{figure}[!htb]
    \centering
    \includegraphics[width=0.7\linewidth]{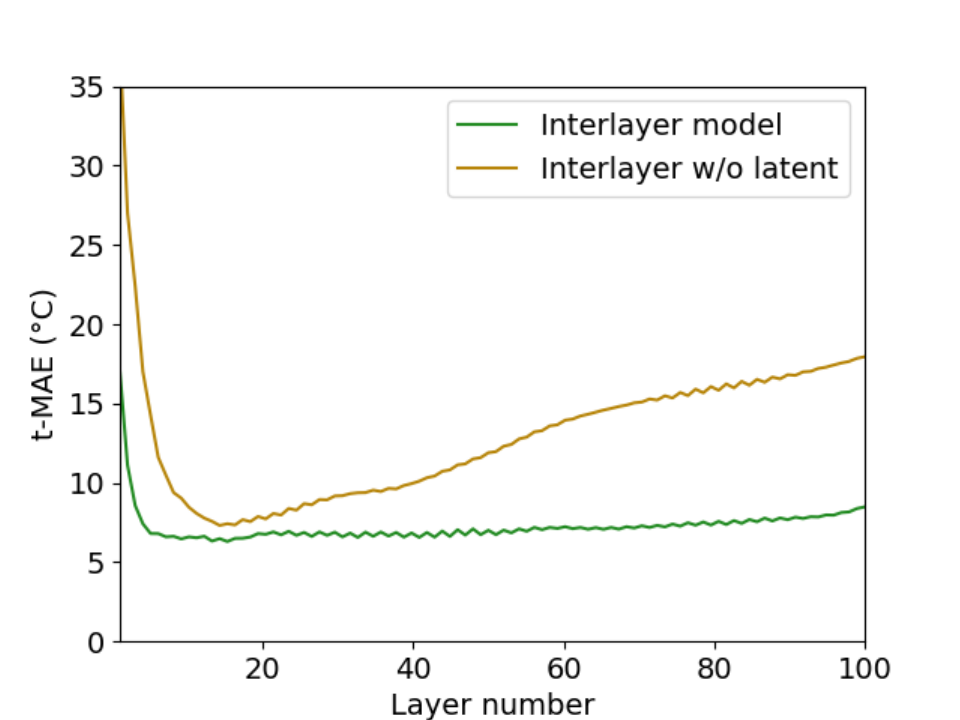}
    \caption{Temporal MAE (t-MAE) on the test set for the interlayer model, compared with its full-dimensional (non-latent) counterpart.}
    \label{fig:tmae-interlayer}
\end{figure}

\begin{figure}[!htb]
    \centering
    \includegraphics[width=0.7\linewidth]{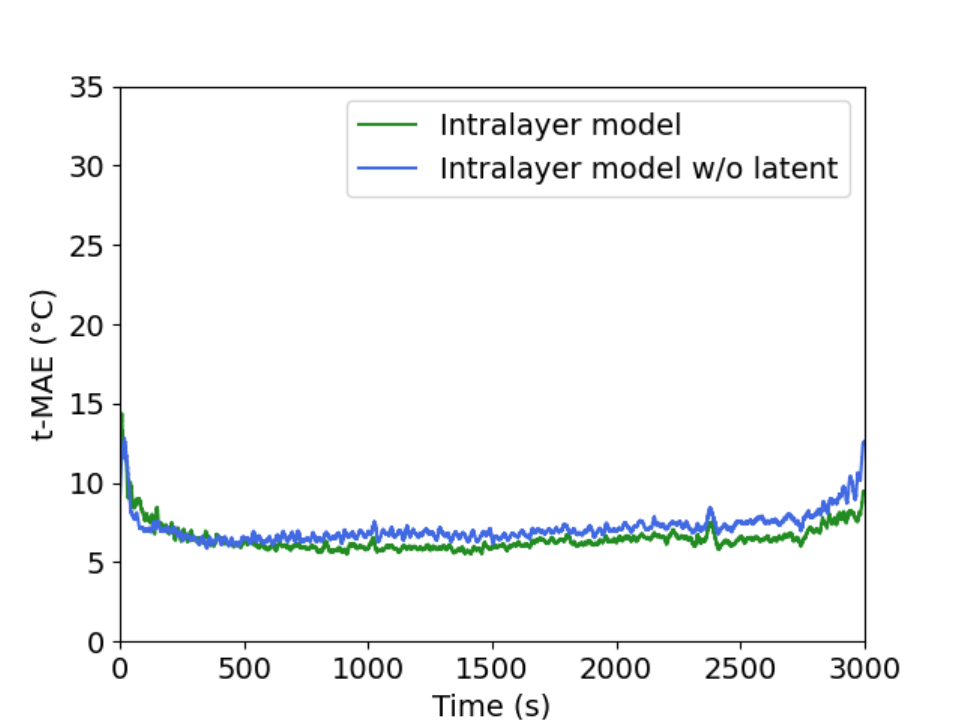}
    \caption{Temporal MAE (t-MAE) on the test set for the intralayer model, compared with its ablated variants.}
    \label{fig:tmae-intralayer}
\end{figure}

\begin{figure}[!htb]
    \centering
    \includegraphics[width=0.7\linewidth]{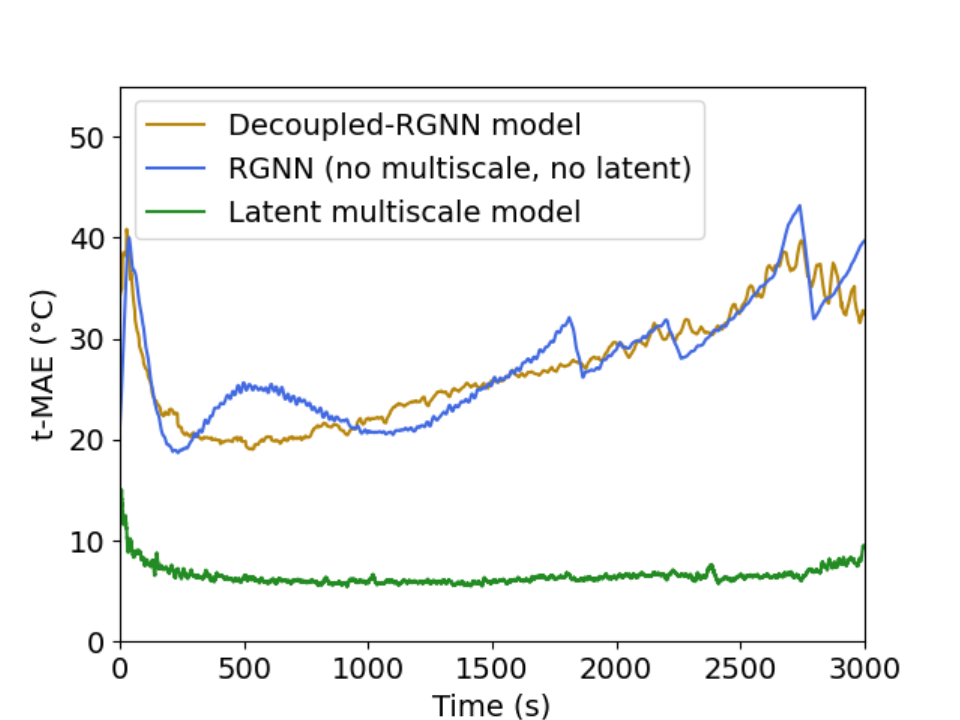}
    \caption{Temporal MAE on the test set for the LM-RGNN model, compared with its ablated variant and the Decoupled-RGNN baseline.}
    \label{fig:tmae-final}
\end{figure}

\paragraph{Temperature Prediction Accuracy.}
Tables~\ref{tab:models-perfs} and \ref{tab:models-perfs2} report the performance metrics of the interlayer, intralayer, and LM-RGNN models, together with their ablated variants and the Decoupled-RGNN baseline. A detailed analysis of the differences between each model and its ablated counterpart is provided in Section~\ref{sec:ablation}. Figures~\ref{fig:temperature-array} and \ref{fig:temperature-array2} in Appendix~\ref{appendix:temperature-figd} provide a visual comparison of temperature predictions, for a representative geometry and time step, between the LM-RGNN model and the Decoupled-RGNN baseline.
Overall, the LM-RGNN (which combines the predictions of the interlayer and intralayer models) achieves consistently better performance than the Decoupled-RGNN baseline. Improvements are observed across temperature prediction accuracy, spatial and temporal gradient errors, and melt-pool estimation. These gains are obtained with a memory footprint comparable to the baseline, while inference time is only marginally higher (+0.97ms/time step).
The interlayer and intralayer models exhibit similar MAE values; however, their MAPE scores differ substantially. 
This difference is primarily due to the lower temperature magnitudes observed in interlayer data, which correspond to end-of-layer cooling phases, as opposed to intralayer sequences that include active laser heating and substantially higher temperatures. As a result, relative errors are magnified for the interlayer model. In contrast, the interlayer model attains lower MME values, reflecting the absence of extreme temperature peaks during cooling periods.
We further note that the VGAE was trained on the full set of temperature fields rather than being specialized for interlayer states, which represent only a small fraction of the dataset. Training a dedicated VGAE on interlayer temperature fields could therefore further improve interlayer prediction accuracy. Finally, although the interlayer model exhibits a higher per-step inference time than the intralayer model due to its larger number of parameters, this has a limited impact on the overall inference cost of the LM-RGNN. Indeed, the interlayer model is invoked only once per printed layer (100 time steps in total), whereas the intralayer model accounts for the majority of time-step predictions.
\paragraph{Temporal Stability.}
Figure~\ref{fig:tmae-final} reports the t-MAE for the LM-RGNN and the Decoupled-RGNN baseline. The LM-RGNN exhibits strong temporal stability, characterized by low-amplitude oscillations and no noticeable error accumulation over time. In contrast, the Decoupled-RGNN baseline shows a gradual but persistent increase in error, together with more pronounced temporal fluctuations.
Further insight is provided in Figures~\ref{fig:tmae-interlayer} and \ref{fig:tmae-intralayer}, which demonstrate the temporal stability of the interlayer and intralayer models, respectively. For the intralayer setting, larger error oscillations are observed at late time steps. This behavior can be partially explained by data sparsity, which affects the reliability of the averaged error estimates, as only a limited number of simulations extend beyond approximately 2800s.
Consistent with the behavior observed for the VGAE in Figure~\ref{fig:tmae-vgae}, an initial error peak is present at early time steps. This effect arises from the computation of MAE over a temporally varying active domain and reflects error concentration during the early stages of the printing process. A detailed analysis of this phenomenon is provided in the Appendix \ref{appendix:tmae-peak}.
\begin{table*}[ht]
\centering
\scriptsize  
\caption{Comparison of performance metrics on the test set across the proposed models, their ablated variants (marked by the $\hookrightarrow$ symbol) , and the baseline.} \label{tab:models-perfs}
\begin{tabular}{lcccccccc} 
    \hline
    Model & MAE (°C) & MAPE (\%) & MME (°C) & mpIoU & $\text{MAE}_{\nabla_{xy}}$ (°C/mm) & $\text{MAE}_{\nabla_{t}}$ (°C/s) & Parameters\\
    \hline
    Interlayer & 7.29 $\pm$ 0.17 & 4.48 $\pm$ 0.22 & 44.5 $\pm$ 0.5 &  & 2.10 $\pm$ 0.05 & 0.76 $\pm$ 0.09 & 85k \\
    $\hookrightarrow$ w/o latent & 17.80 $\pm$ 9.16 & 4.73 $\pm$ 1.57 & 95.0 $\pm$ 40.6 &  & 4.65 $\pm$ 1.54 & 0.40 $\pm$  0.18 & 91k \\
    \hline
    Intralayer & 6.48 $\pm$ 0.19 & 2.20 $\pm$ 0.010 & 108.7 $\pm$ 2.80 & 0.87 $\pm$ 0.0016 & 2.46 $\pm$ 0.030 & 1.61 $\pm$ 0.020 & 42k \\  
    $\hookrightarrow$ w/o latent & 6.80 $\pm$ 0.34 & 2.43 $\pm$ 0.18 & 50.6 $\pm$ 2.0 & 0.91 $\pm$ 0.0081 & 2.27 $\pm$ 0.10 & 1.28 $\pm$ 0.064 & 41k \\  
    \hline
    LM-RGNN & 6.50 $\pm$ 0.10 & 2.58 $\pm$ 0.002 & 108.7 $\pm$ 0.28 & 0.85 $\pm$ 0.0017 & 2.44 $\pm$ 0.030 & 1.98 $\pm$ 0.0031 & 127k \\ 
    $\hookrightarrow$ w/o latent \& multiscale strategy & 26.28 $\pm$ 1.45 & 7.25 $\pm$ 1.12 & 111.46 $\pm$ 5.1 & 0.69 $\pm$ 0.097 & 5.98 $\pm$ 1.30 & 7.13 $\pm$ 0.84 & 131k \\ 
    \hline
    Baseline Decoupled-RGNN & 26.57 $\pm$ 3.44 & 6.59 $\pm$ 1.26 & 170.9 $\pm$ 19.5 & 0.67 $\pm$ 0.032 & 6.41 $\pm$ 1.28 & 7.79 $\pm$ 0.70 & 111k \\ 
    \hline
\end{tabular}
\end{table*}
\begin{table}[t]
\centering
\scriptsize  
\caption{Comparison of per-time-step inference time and training-time memory pressure across the proposed models, their ablated variants (marked by the $\hookrightarrow$ symbol), and the baseline.} \label{tab:models-perfs2}
\begin{tabular}{lcc}
    \hline
    Model & Inference time (ms) & Mem. pressure \\
    \hline
    Interlayer & 30.4 & \textit{Low} \\
    $\hookrightarrow$ w/o latent & 30.9 & \textit{Medium} \\
    \hline
    Intralayer & 12.35  & \textit{Low} \\  
    $\hookrightarrow$ w/o latent & 14.28  & \textit{Medium} \\  
    \hline
    LM-RGNN & 14.47  & \textit{Low} \\ 
    $\hookrightarrow$ w/o latent \& multiscale strategy & 68.19  & \textit{High} \\ 
    \hline
    Baseline Decoupled-RGNN & 13.50 & \textit{Low} \\ 
    \hline
\end{tabular}
\end{table}


\subsection{Ablation Study} \label{sec:ablation}

Tables~\ref{tab:models-perfs} and \ref{tab:models-perfs2}, together with Figures~\ref{fig:tmae-interlayer}, \ref{fig:tmae-intralayer}, and \ref{fig:tmae-final}, report performance metrics and temporal MAE results for the interlayer, intralayer, and LM-RGNN models, as well as their corresponding ablated variants.
For the interlayer setting, we compare the latent-RGNN architecture with a non-latent ablation that operates directly on full-resolution temperature fields while maintaining a comparable model size. The non-latent variant exhibits substantially higher training variability, increased memory consumption, and degraded performance across most metrics, with the exception of $\text{MAE}_{\nabla_{t}}$, which remains low for both variants.
In the intralayer setting, the latent-RGNN and its non-latent counterpart achieve comparable performance. The non-latent model yields slightly improved MME, melt-pool estimation, and gradient-related metrics; however, these gains come at the cost of higher inference time and significantly increased memory usage due to full-field processing. Consequently, for the intralayer model, operating in the latent space primarily provides benefits in terms of memory efficiency rather than predictive accuracy.
Overall, latent representations yield more pronounced performance gains for the interlayer model than for the intralayer model. We assume that this difference arises from the nature of the temporal dynamics: consecutive intralayer states are highly similar, with changes largely confined to local regions around the laser path, whereas interlayer states differ more substantially due to layer addition, extended thermal diffusion, and cumulative energy input. In this context, predicting intralayer transitions is well suited to message-passing architectures with limited receptive fields. By contrast, the latent representation, defined on a coarser mesh, enables more efficient global information propagation for interlayer dynamics, followed by accurate reconstruction via the pretrained VGAE decoder. 
Finally, we compare the LM-RGNN with an ablated non-latent RGNN trained on full sequences without temporal multiscale decomposition. The ablated model performs significantly worse. One contributing factor is the difficulty of training RGNNs on extremely long sequences (up to 1800 time steps after subsampling), which leads to poor gradient propagation, unstable optimization, and oscillatory training loss. In addition, operating on full-resolution temperature fields substantially increases memory requirements, forcing the use of very small batch sizes and further degrading training stability.

%% file: Conclusion.tex
\section{Conclusion}
\paragraph{Contributions.}
We introduced a LM-RGNN framework for predicting full temperature histories in additive manufacturing directly on the part mesh, conditioned on geometric and process parameters. The integration of a VGAE with a temporal multiscale RGNN enables stable and accurate long-horizon prediction while mitigating the memory burden associated with high-resolution simulation data.
The explicit decomposition into interlayer and intralayer models allows the framework to capture both slow, diffusion-dominated thermal evolution and fast, laser-induced spatially localized dynamics, leading to improved accuracy of predicted temperature fields, spatial and temporal gradients, and melt-pool localization relative to existing baseline.
Extensive experiments on simulated powder bed fusion data demonstrate that the proposed architecture achieves accurate and temporally stable predictions over thousands of time steps across diverse geometries. 
\paragraph{Limitations and Future Work.}
While the proposed framework is designed with three-dimensional geometries and broader process generalization in mind, the present study is limited to two-dimensional simulations and a single powder bed fusion process with fixed parameters. Moreover, the models are trained exclusively on simulated data, and further investigation is needed to assess robustness to process variability and transferability to experimental measurements.
\paragraph{Broader Impact.}
Beyond additive manufacturing, the proposed latent-multiscale-RGNN framework provides a general approach for modeling physics-driven systems with coupled fast and slow temporal dynamics on complex geometries. Potential application domains include fluid dynamics and climate and weather modeling
(see Appendix \ref{appendix:general-appli} for a detailed discussion).

%% file: Appendix.tex
\section{General Applicability of the Proposed Framework} \label{appendix:general-appli}
The proposed LM-RGNN framework targets applications involving spatiotemporal physical fields defined on meshes- (potentially at high spatial resolution to capture localized phenomena) that exhibit pronounced temporal multiscale behavior arising either from the underlying physics or from time-varying inputs such as boundary conditions or external sources. The framework naturally accommodates variable geometries with geometry-specific meshes and, through masking, supports progressively activated spatial domains over time.
Representative application domains for which the LM-RGNN architecture could be considered include, but are not limited to:
\begin{itemize}
\item Hemodynamic flow modeling in coronary arteries with variable geometries, such as in the presence of lesions, is characterized by pronounced temporal multiscale behavior driven by the cardiac cycle \cite{candreva2022current}. Within each heartbeat, fast pulsatile dynamics (systolic and diastolic pressure/flow waves) drive rapid changes in velocity, pressure, and wall shear stress. These fluctuations occur on the order of fractions of a second and are crucial to hemodynamic function and pathology (e.g., atherogenesis) \cite{boutsianis2004computational}. Over longer periods, variations across cardiac cycles (e.g., changes in heart rate, vessel compliance, stenosis severity) modulate amplitude and phase relationships of flow features and have been investigated as potential slow temporal biomarkers \cite{tanade2024establishing}.
The temporal multiscale structure of coronary hemodynamics makes this domain a natural candidate for multiscale sequence modeling. The LM-RGNN framework is well aligned with this setting, as its architecture explicitly separates fast local dynamics from slower global evolution, which may help improve the stability and accuracy of long-horizon predictions without assuming strict periodicity. \\
High-fidelity CFD simulations on patient-specific coronary geometries are an important tool for personalized hemodynamic analysis and clinical decision support \cite{psiuk2024methodology}. However, these simulations are computationally intensive and difficult to run in real time or over long sequences of cardiac cycles. Learning-based surrogate models that generalize across geometries and operate directly on unstructured meshes offer a promising alternative. In this context, the LM-RGNN framework could be used to model flow evolution directly on patient-specific artery meshes while maintaining temporal consistency over extended sequences. \\
Several machine learning approaches have recently been explored for coronary hemodynamic prediction, including deep vectorized operators \cite{suk2025deep}, voxel-based methods such as Voxel2Hemodynamics \cite{ni2023voxel2hemodynamics}, and mesh neural networks \cite{suk2024mesh}. These works demonstrate the feasibility of data-driven hemodynamic modeling and further motivate the exploration of temporally multiscale, mesh-based recurrent frameworks.
\item Weather and climate modeling on unstructured meshes involves the simulation of physical fields such as temperature, pressure, wind, and humidity that evolve across multiple temporal scales. Fast processes, including diurnal cycles, convection, and frontal systems, typically unfold over hours to days, whereas slower phenomena such as seasonal variability, interannual modes, and long-term climate trends evolve over months to decades. This pronounced temporal multiscale structure presents a challenge for long-horizon prediction and requires modeling approaches that can simultaneously represent short-term variability and longer-term evolution \cite{de2023machine}.
By separating the temporal modeling into fast and slow components, LM-RGNN can model rapid weather dynamics and slower climate trends in a manner similar to how its interlayer/intralayer decomposition isolates fast and slow thermal phenomena in additive manufacturing.\\
Contemporary numerical weather and climate models increasingly rely on variable-resolution or unstructured meshes to adapt spatial resolution to regions of interest, such as storms or complex topography, while maintaining computational efficiency elsewhere \cite{de2023machine}.
Graph-based learning approaches can operate directly on such meshes, avoiding interpolation to regular grids and helping preserve geometric fidelity over long forecasts. In this context, the LM-RGNN architecture, which operates natively on simulation meshes and leverages a variational graph autoencoder to learn compact latent representations on coarser graphs, offers a potentially efficient means of modeling spatiotemporal dynamics while controlling computational cost.\\
Recent advances in machine learning for weather and climate forecasting further highlight the importance of flexible spatial representations and long-range temporal modeling, as illustrated by approaches such as GraphCast \cite{lam2023learning} and ClimaX \cite{nguyen2023climax}.
\item Unsteady flow in turbomachinery compressors results from interactions between rotating and stationary components, such as blade rows, leading to periodic fluctuations in pressure, velocity, and turbulence intensity. Steady-flow assumptions commonly adopted in compressor design may be insufficient to capture these effects, particularly in the presence of rotor–stator interactions, shock dynamics, or stall inception, motivating the use of explicitly unsteady analyses. Moreover, the coupling between fast blade-passing phenomena and slower variations in operating conditions (e.g., off-design operation or transient loading) gives rise to temporal multiscale behavior in compressor flows \cite{holmes2011unsteady}. This makes long-horizon prediction of unsteady flow a challenging task for surrogate models. Within this context, the LM-RGNN framework provides a structured approach for jointly representing fast oscillatory dynamics and slower temporal trends directly on mesh-based discretizations. 
\end{itemize}
\section{Details about the Dataset} \label{appendix:dataset-details}
\subsection{Geometry Parameterization}\label{appendix:subsec:geo-param}
\begin{figure}[!htb]
    \centering
    \includegraphics[width=0.25\linewidth]{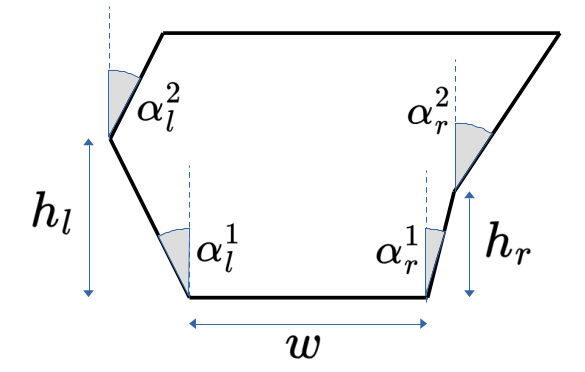}
    \caption{Geometry parametrization. $w$ is the part base width, $h_l$ and $h_r$ are the heights of left and right line-breaks respectively, $\alpha_l^1$, $\alpha_r^1$, $\alpha_l^2$, $\alpha_r^2$ are line angles.}
    \label{fig:geomparam}
\end{figure}

\begin{figure}[!htb]
    \centering
    \begin{subfigure}[b]{\linewidth}
        \centering
        \includegraphics[width=0.55\linewidth]{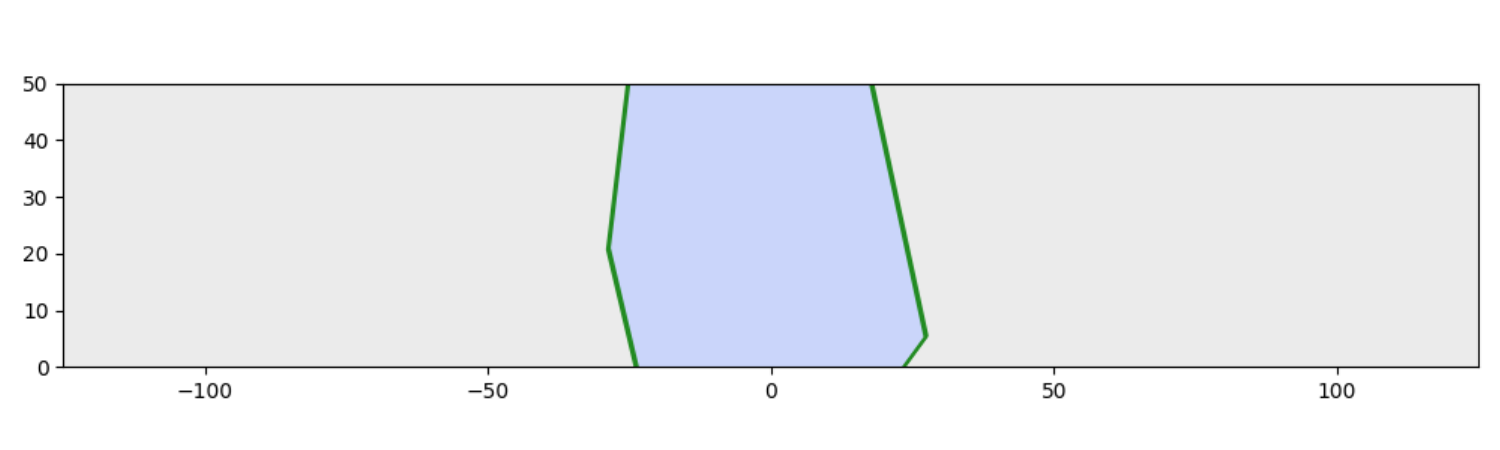}
    \end{subfigure}\hfill
    \begin{subfigure}[b]{\linewidth}
        \centering
        \includegraphics[width=0.55\linewidth]{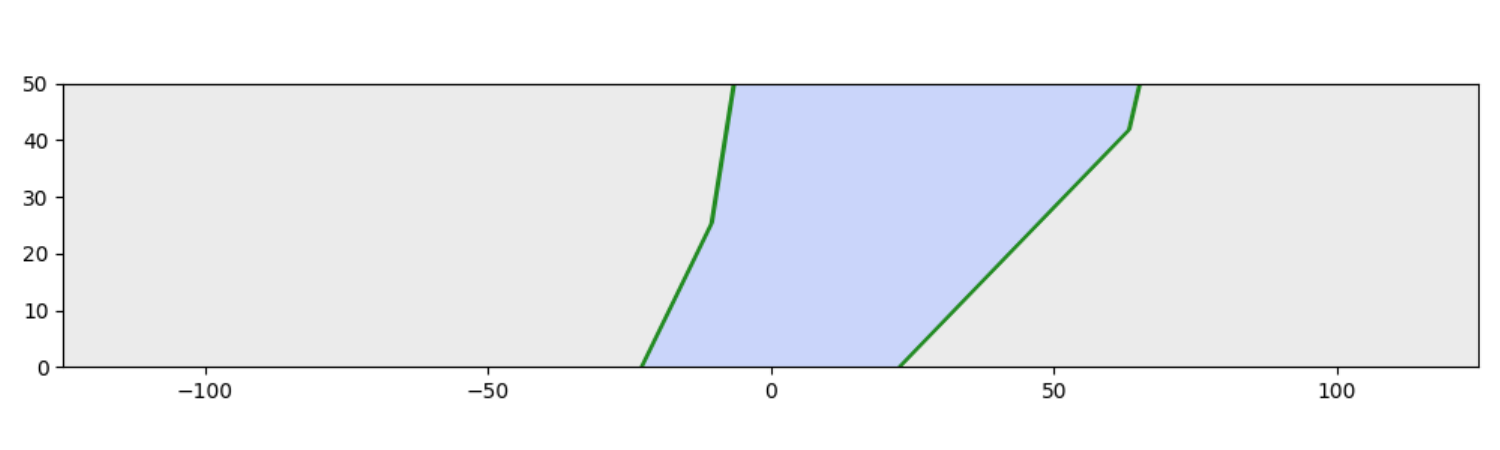}
    \end{subfigure}\hfill
    \begin{subfigure}[b]{\linewidth}
        \centering
        \includegraphics[width=0.55\linewidth]{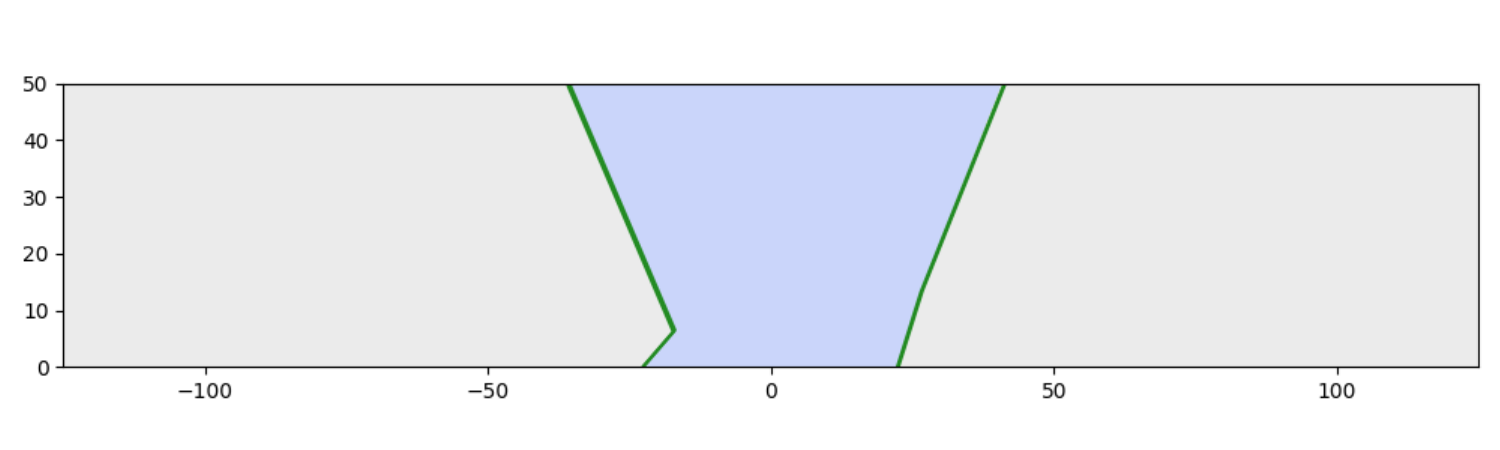}
    \end{subfigure}
    \caption{Examples of generated geometry. Metallic powder is in gray, the final printed part is in blue and the part boundaries are highlighted in green.}
    \label{fig:geom_examples}
\end{figure}

In the powder bed fusion process, the part is built layer by layer from spread metallic powder that is selectively melted. Upon completion of the process, the fully consolidated metallic part remains embedded within the surrounding powder bed. Accordingly, the simulation domain includes both the solid metal region and the adjacent powder material as shown in Figure \ref{fig:geom_examples}.
Two-dimensional solid part geometries are parameterized using seven design variables (see Figure \ref{fig:geomparam}) that control both shape and size. The overall part height is fixed at 50mm and sliced into 100 layers of thickness 0.5mm. The parameterized geometries are defined by two piecewise-linear side boundaries, each consisting of two segments that may have different slopes. The shape of each side part boundary is individually controlled by three geometric parameters: one height and two angles, while the part base width is defined by an additional geometric parameter. Angle parameters can vary within the range [$\pi/4$; $3\pi/4$ ] and the base width ranges from $10mm$ to $90mm$. The lower bound of angle range represents a key manufacturability constraint imposed to prevent impractical layer superposition during the build process.
The parametrization results in variable part widths with distinct start and end points in each layer. The overall variability in the geometry contributes to a diverse range of thermal responses.

\subsection{Dataset Generation}\label{appendix:subsec:dataset-desc}
The modeling of the powder bed fusion process is restricted to a two-dimensional thermal framework based on a layer-by-layer approach employing a mesh element activation algorithm. The finite element mesh discretizes all layers deposited on the base plate. To preserve quadrilateral mesh elements, the number of elements is kept constant for each layer of both the solid part and the surrounding powder region; this number is determined from the average width of all layers composing the part. Consequently, the mesh size remains constant along each layer. However, across the powder region, the mesh size is gradually coarsened away from the part, enabling a smooth transition between material domains and reducing numerical errors associated with abrupt changes in thermal properties when the heat source approaches layer boundaries.

The transient thermal analysis is discretized into 100 layer-deposition steps using a mesh element activation strategy. For each step, laser heat input is applied over a selectively activated region through a moving heat source. The laser power is maintained at $175 W$ and translated at a constant scanning velocity of $5 mm/s$ during the exposure phase. Upon completion of the laser scan for each layer, a cooling phase corresponding to a recoating time of $10 s$ is simulated before activation of the subsequent layer. The scan path is restricted to the solid region of each layer and alternates in direction between successive layers.

\begin{table}[ht]
\centering
\small
\caption{Simulations statistics}
\label{annexe:summary_stats}
\begin{tabular}{lccc}
\toprule
 & Number of elements & Runtime & Number of time steps \\
\midrule
mean & 29987 & 1h 38m 37s & 9316 \\
std\_dev & 5756 & 0h 53m 13s & 1341 \\
min & 18126 & 0h 20m 42s & 6544 \\
max & 43670 & 3h 51m 2s & 12523 \\
\bottomrule
\end{tabular}
\end{table}

The transient thermal analysis is performed using a finite element solver based on Euler implicit time-integration scheme.
The dataset comprises 140 parametric combinations generated using a Latin-hypercube-based algorithm (Latin Centroidal Voronoi Tessellation (LCVT), see \cite{saka2007latinized}) to sample the seven design parameters. All simulations were executed on a high-performance computing cluster using 10 SMP processors based on the AMD EPYC Zen 3 (Milan) architecture, with 6 GB of memory allocated per simulation. A summary of key statistics is provided in Table \ref{annexe:summary_stats}, including the number of mesh elements, the number of time steps, and the computational runtime. The number of mesh elements reflects the proportion of the solid part relative to the total domain covering the 100 layers. The number of time steps is proportional to the cumulative width of the solid regions across layers, corresponding to the total laser scan path length. The average runtime per simulation in the dataset is approximately 1.5 hours.
\subsection{Thermal Simulation Model}\label{appendix:subsec:sim-model}
\begin{figure}[htb]
    \centering
    \includegraphics[width=0.7\linewidth]{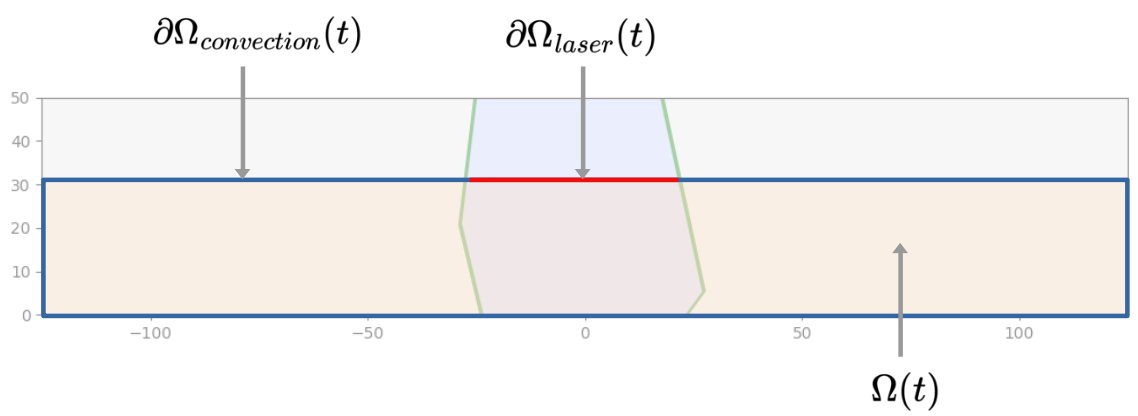}
    \caption{Thermal domain at time $t$.}
    \label{appendix:fig:thermal-domain}
\end{figure}

\begin{figure*}[htb]
    \centering
    \begin{subfigure}[b]{0.475\textwidth}
        \centering
        \includegraphics[width=0.75\textwidth]{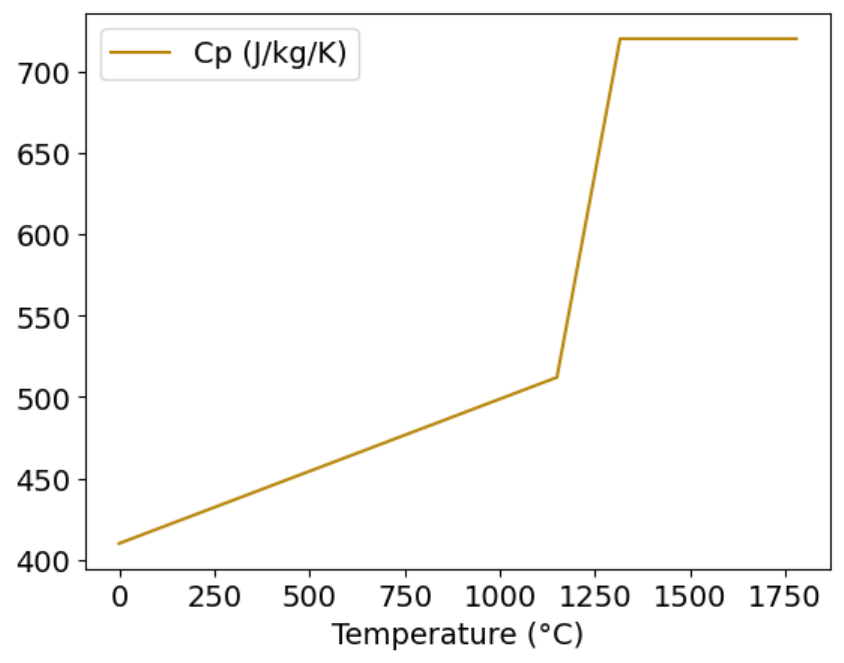}
        \caption{\small Evolution of heat capacity $C_p$ with temperature.}  
        \label{appendix:fig:coeff-cp}
    \end{subfigure}
    \hfill
    \begin{subfigure}[b]{0.475\textwidth}  
        \centering 
        \includegraphics[width=0.75\textwidth]{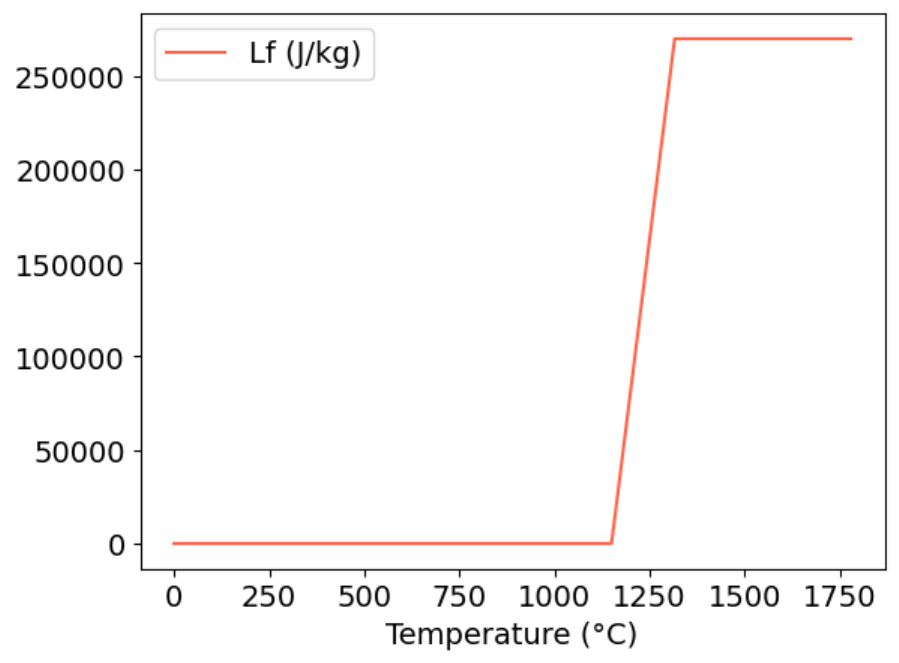}
        \caption{\small Evolution of latent heat $L_f$ with temperature.}
        \label{appendix:fig:coeff-lf}
    \end{subfigure}
    \vskip\baselineskip
    \begin{subfigure}[b]{0.475\textwidth}   
        \centering 
        \includegraphics[width=0.75\textwidth]{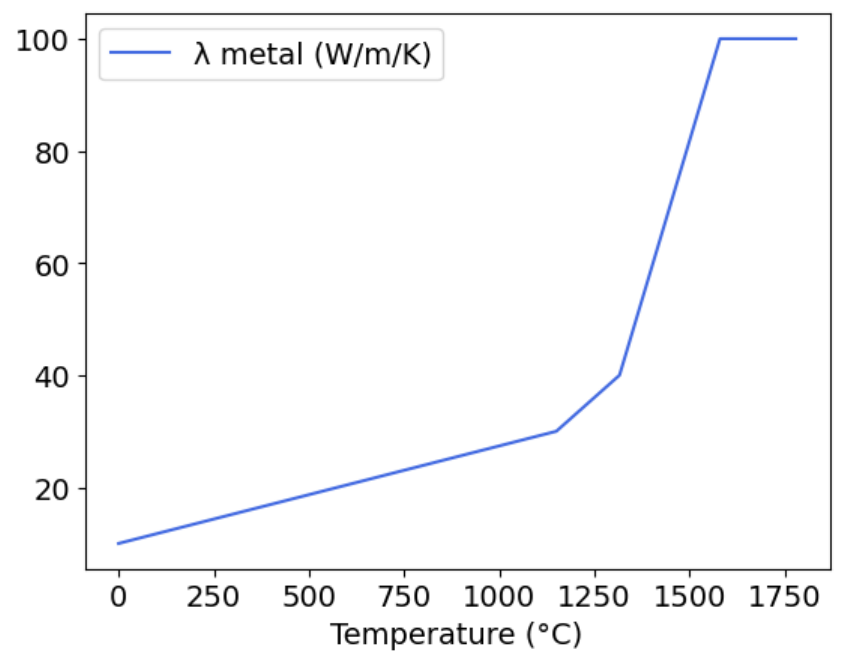}
        \caption{\small Evolution of heat conductivity of the metal $\lambda_{metal}$ with temperature.}
        \label{appendix:fig:coeff-lambda-metal}
    \end{subfigure}
    \hfill
    \begin{subfigure}[b]{0.475\textwidth}   
        \centering 
        \includegraphics[width=0.75\textwidth]{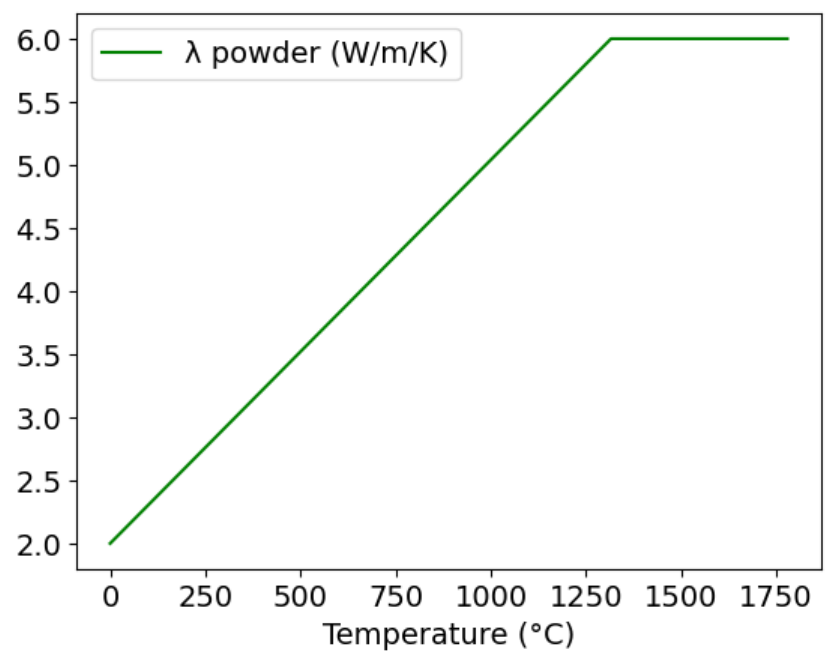}
        \caption{\small Evolution of heat conductivity of the metallic powder $\lambda_{powder}$ with temperature.}
        \label{appendix:fig:coeff-lambda-powder}
    \end{subfigure}
    \caption{\small Thermo-physical properties used in the numerical model.} 
    \label{appendix:fig:coeffs}
\end{figure*}

The thermal evolution is governed by the transient heat equation with phase-change effects of a specific metallic material,
\begin{equation*}
\begin{aligned}
     C_p(T) \rho \frac{\partial T}{\partial t} + \rho \frac{\partial L_f(T)}{\partial t} &= \nabla .(\lambda(T) \nabla T), \;\; \forall (x,y) \in \Omega(t), \text{ with }T=T(x,y,t), \\ 
     - \lambda(T) \nabla T.\vec{n} &= h (T - T_{ext}), \;\; \forall (x, y) \in \partial \Omega_{convection}, \\
     \lambda(T) \nabla T.\vec{n} &= Q(x,t), \;\; \forall (x, y) \in \partial \Omega_{laser},
\end{aligned}
\end{equation*}
where $T$ denotes the temperature, $\rho$ is the material density, $C_p(T)$ the temperature-dependent heat capacity, $\lambda(T)$ the thermal conductivity, and $L_f(T)$ the latent heat of fusion.
The laser heat input is imposed through a Neumann boundary condition,
\begin{equation*}
Q(x,t) = S(t) \frac{3P}{\pi r^2} \text{exp}\left(- \frac{(x-x_l(t))}{r^2}\right),
\end{equation*}
with laser power $P=175\ W$, beam radius $r=2\ mm$, and $x_l(t)$ the laser position moving at constant speed along the prescribed print path. The switching function $S(t)$ equals 1 during heating phases and 0 during cooling phases.
The computational domain $\Omega(t)$ evolves over time to account for layer deposition through element activation strategy; the top boundary is incrementally displaced upward by $0.5\ mm$ with the addition of each new layer. The laser boundary $\partial \Omega_{laser}$ corresponds to the printed region of the top surface, while the remaining boundary $\partial \Omega_{convection} = \partial \Omega(t) \setminus \partial \Omega_{laser}$ is subject to convective heat transfer with coefficient $h = 10 \ W.m^{-2}$ and external temperature $T_{ext} = 20$°C (see Figure \ref{appendix:fig:thermal-domain}).
The material density is fixed to $\rho = 8190 \ kg.m^{-3}$ (Inconel~718 material). The thermal conductivity $\lambda(T)$ takes distinct values in the powder and solidified metal regions, denoted $\lambda_{powder}(T)$ and $\lambda_{metal}(T)$, respectively.
Temperature-dependent material properties, including $Cp$, $\lambda_{metal}$, $\lambda_{powder}$ and $L_f$ are specified in Figure \ref{appendix:fig:coeffs}.
\section{LM-RGNN Architecture Details} \label{appendix:arch-details}
\subsection{Motivation for Full-Sequence Recurrent Modeling over Autoregressive Approaches} \label{appendix:sec:RNN-motivations}
The output of our simulations consists of time-resolved sequences of temperature fields, and the objective of this work is to learn a mapping from the process parameters (specifically the laser path) to these temperature sequences. A common approach for such temporal prediction tasks is to employ an autoregressive architecture, in which each temperature field is predicted from previous fields and the evolving process parameters. However, autoregressive models are known to suffer from error accumulation over time and often exhibit degraded performance when predicting long sequences, typically beyond a few hundreds time steps (see, e.g., \cite{yue2024tgn}). Several methods have been proposed to mitigate this issue, including techniques such as PDE-based refiners \cite{lippe2023pde} and related stabilization approaches. Unlike recurrent models trained on full-length sequences, autoregressive architectures reduce the depth of the trained neural network by operating on shorter subsequences. Nevertheless, our objective is to accurately predict temperature fields over very long temporal horizons and to generalize to even longer sequences corresponding to larger parts, potentially involving tens of thousands of time steps. This requirement necessitates a model with strong temporal stability and minimal error accumulation. A natural solution is to train a recurrent neural network on full-length sequences. Because its hidden state acting as a dynamical memory, an RNN can attenuate the propagation of past prediction errors, effectively behaving as a low-pass filter and re-anchoring predictions through the learned temporal structure.
\subsection{GraphGRU: Graph Gated Recurrent Unit} \label{appendix:subsec:graphGRU}
GRUs were chosen as the basis for the GraphGRU due to their architectural simplicity and favorable gradient propagation properties, as well as their lower memory footprint compared to transformer-based architectures. However, relative to transformers, GRUs exhibit more limited capacity for modeling long-range temporal dependencies.\\
We define a graph $G = (A, V, E, g)$ where $A$ denotes the adjacency matrix, $E$ the associated edge features, $V$ the node features and $g$ the global attributes. When the graph topology $A$ and edge features $E$ are fixed and we use the notation $G[V, g] = (A, V, E, g)$.
We consider a temporal sequence of graphs $\mathcal{G} = \lbrace\mathcal{G}_t\rbrace_{t=1..N} = \lbrace G[V(t), g(t)] = (A, V(t), E, g(t)\rbrace_{t=1..N}$ in which node features $V(t)$ and global attributes $g(t)$ evolve over time while the graph topology remains fixed. We define a GraphGRU, taking as input a node feature input state $X_t$ and a node feature hidden state $H_{t-1}$, by:
\begin{equation}
    \begin{aligned}
        &Z_t = \mathrm{sigmoid}(\mathrm{MP}_z(G[X_t \oplus H_{t-1}, g_t])), \\
        &R_t = \mathrm{sigmoid}(\mathrm{MP}_r(G[X_t \oplus H_{t-1}, g_t])), \\
        &\hat{H}_t = \mathrm{tanh}(\mathrm{MP}_h(X_t, R_t \odot H_{t-1}, g_t)), \\
        &H_t = (1-Z_t) \odot H_{t-1} + Z_t \odot \hat{H}_t, \\
        &\mathrm{GraphGRU}(X_t, H_{t-1}, g_t) = H_t,
    \end{aligned}
\end{equation}
where $\text{MP}_{z}$, $\text{MP}_{r}$ and $\text{MP}_{h}$ are message passing neural networks that take as input a graph $G$ and return an output node features vector. The quantities $X_t$, $H_{t-1}$, $H_t$, $\hat{H}_t$, $Z_t$ and $R_t$ are nodes features vectors. The operator $\oplus$ denotes the node feature-wise concatenation between two node features vectors and the operator $\odot$ denotes the element-wise multiplication between two node features vectors.
\subsection{RGNN Architecture} \label{appendix:sec:RGNN-arch}
\begin{center}
    \includegraphics[width=0.45\linewidth]{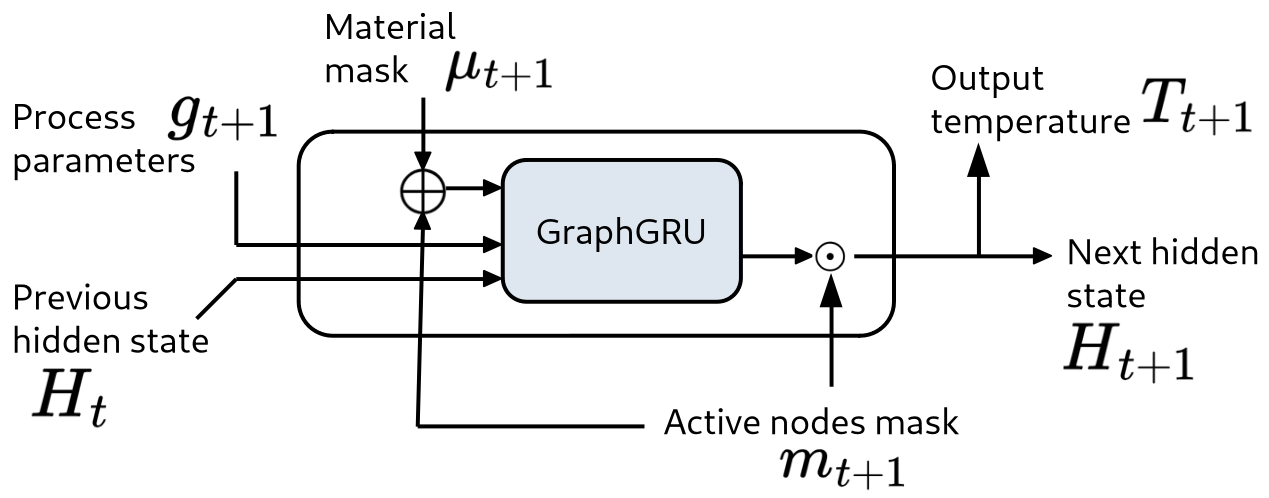}
    \captionof{figure}{RGNN architecture.}
    \label{fig:rgnn}
\end{center}
\subsection{VGAE: Variational Graph Auto-Encoder} \label{appendix:sec:VGAE}
\begin{figure}[!htb]
    \centering
    \begin{subfigure}[t]{0.45\columnwidth}
        \centering
        \includegraphics[width=0.5\linewidth]{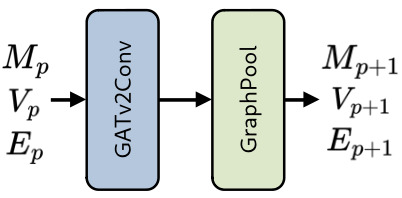}
        \caption{}
    \end{subfigure}%
    \hfill
    \begin{subfigure}[t]{0.45\columnwidth}
        \centering
        \includegraphics[width=0.5\linewidth]{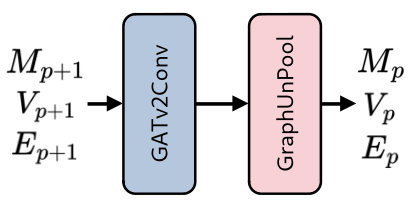}
        \caption{}
    \end{subfigure}%
    \caption{(a) VGAE encoder block. (b) VGAE decoder block.}
    \label{fig:vgae_blocks}
\end{figure}

\begin{figure}[!htb]
    \begin{subfigure}[t]{\columnwidth}
        \centering
        \includegraphics[width=0.35\linewidth]{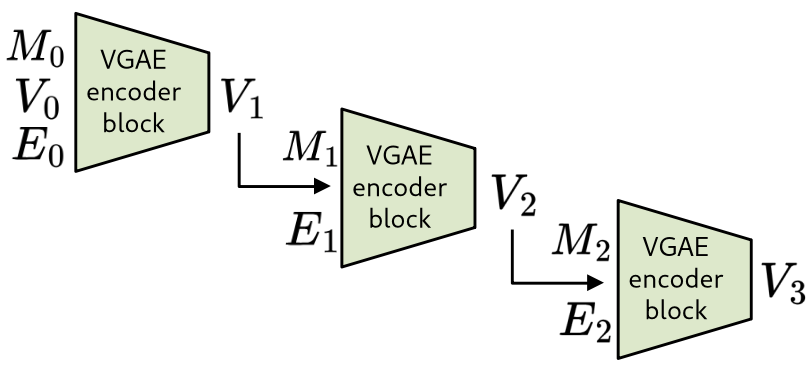}
        \caption{}
    \end{subfigure}%
    
    \medskip
    \begin{subfigure}[t]{\columnwidth}
        \centering
        \includegraphics[width=0.35\linewidth]{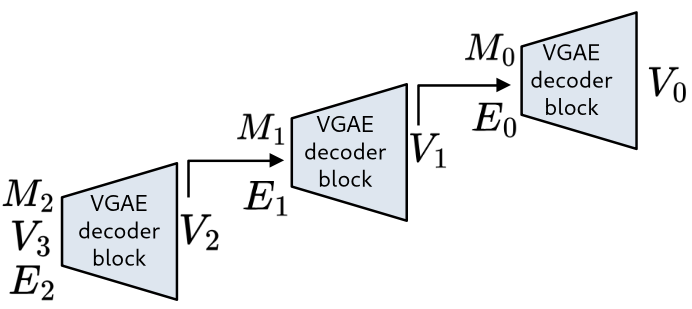}
        \caption{}
    \end{subfigure}%
    \caption{(a) VGAE encoder with depth $d=3$. (b) VGAE decoder with depth $d=3$.}
    \label{fig:vgae_encdec}
\end{figure}

\paragraph{Propagation of Local Information in VGAE and Flat GNNs.}
The VGAE architecture is well suited for capturing multiscale geometric and physical features due to the use of hierarchical graph pooling and unpooling operations. These mechanisms facilitate efficient propagation of local information across the graph hierarchy, in contrast to flat GNN architectures in which information propagation is constrained by network depth. As a result, flat GNNs typically require deeper networks to capture long-range dependencies, increasing model complexity and training difficulty. 
\paragraph{Compressed Latent Mesh.} In contrast to \cite{kipf2016variational}, the adjacency matrix of the compressed graph is constructed deterministically rather than being learned by the VGAE. Due to the rectangular simulation domain and the quadrilateral mesh structure, a hierarchy of sub-meshes can be constructed efficiently. Specifically, given a mesh $M_p$, we construct a coarser sub-mesh $M_{p+1}$ by retaining one node out of two while always preserving boundary nodes and nodes located at the metal-powder interface. For each edge in the sub-mesh $M_{p+1}$, we associate a feature vector encoding the Euclidean distance and unit direction vector between the corresponding nodes, yielding the edge-feature set $E_{p+1}$. By recursively applying this procedure to an initial mesh $M_0$, we deterministically construct a sequence of nested meshes $\lbrace M_p \rbrace_{p=0}^L$ and their associated edge-feature sets $\lbrace E_p \rbrace_{p=0}^L$.
\paragraph{Graph Pooling.} Let $M_p$ denote a mesh and $M_{p+1}$ the corresponding sub-mesh constructed as described above. We define a graph pooling operator that maps node features from $M_p$ to $M_{p+1}$. For each node $n_i \in M_{p+1}$, we consider the set $P_i = \lbrace n_j \in M_p \mid n_j \text{ is a neighbor of } n_i \text{ in } M_p \rbrace$. The pooled node feature $v_i$ is then computed as $v_i = \sum_{n_j \in P_i} \mathrm{ MLP}(\mathrm{LayerNorm}(e_{n_j \rightarrow n_i} \oplus u_j))$, where $e_{n_j \rightarrow n_i}$ denotes the edge feature in $M_p$ between the nodes $n_j$ and $n_i$, $u_j$ is the node-feature of $n_j$ and $\oplus$ denotes feature concatenation. This pooling operation defines a mapping $(M_p, V_p, E_p) \rightarrow (M_{p+1}, V_{p+1}, E_{p+1})$, where $V_{p+1} = \lbrace v_i \text{ for } n_i \in M_{p+1}\rbrace$, and $E_{p+1}$ are edge features on $M_{p+1}$.
\paragraph{Graph Unpooling.} Conversely, we define a graph unpooling operator that maps node features from $M_{p+1}$ back to $M_p$. For each node $n_i \in M_p$, we identify the closest node $n_j \in M_{p+1}$ based on their coordinates. The unpooled node feature $u_i$ is computed as $u_i = \mathrm{MLP}(\mathrm{LayerNorm}(e_{n_j \rightarrow n_i} \oplus v_j))$, where $e_{n_j \rightarrow n_i}$ denotes the edge feature of $M_p$ between the nodes $n_j$ and $n_i$, $v_j$ is the node-feature of $n_j$ in $M_{p+1}$. This defines the mapping $(M_{p+1}, V_{p+1}, E_{p+1}) \rightarrow (M_p, V_p, E_p)$, where the set of unpooled node features $V_{p} = \lbrace u_i \text{ for } n_i \in M_{p}\rbrace$.
\paragraph{VGAE Encoder \& Decoder.} We define the building blocks of the VGAE encoder and decoder as
\begin{equation}
    \begin{aligned}
        &\mathrm{EncoderBlock} = \mathrm{GATv2Conv} \circ \mathrm{GraphPool}, \\
        &\mathrm{DecoderBlock} = \mathrm{GATv2Conv} \circ \mathrm{GraphUnPool}, \\
    \end{aligned}
\end{equation} 
as illustrated in Figure \ref{fig:vgae_blocks}, where $\mathrm{GATv2Conv}$ denotes the graph attention convolution introduced by \cite{brody2021attentive}. Let $d$ denote the depth of the VGAE. The encoder and decoder are then constructed by stacking $d$ successive $\mathrm{EncoderBlock}$ and $\mathrm{DecoderBlock}$ modules respectively (see Figure \ref{fig:vgae_encdec}).
\paragraph{VGAE Architecture \& Loss.} The VGAE consists of an encoder that maps an input field to a node-wise latent distribution parameterized by mean and variance vectors, denoted by $\mu$ and $\sigma$, respectively. Following the standard variational autoencoder framework, latent variables are sampled as $z \sim \mathcal{N}(\mu, \sigma)$ and subsequently passed through the VGAE decoder to reconstruct the input field (see Figure \ref{fig:framework}).
The VGAE is trained using a loss function adapted from the $\beta$-VAE formulation (see \cite{higgins2017beta}):
\begin{equation}
    \begin{aligned}
        &\mathcal{L} = \mathcal{L}_{recon} + \beta \mathcal{L}_{KL}, \\
        &\mathcal{L}_{recon} = \frac{1}{N} \sum_{b=1}^N \frac{1}{|m_b|} \sum_{i \in V_0^b} m_{b,i} . || T_{b,i}^{true} - T_{b,i}^{recon} ||^2, \\
        &\mathcal{L}_{KL} = - \frac{1}{N} \sum_{b=1}^N  \frac{1}{2} \sum_{i \in V_d^b} (1 + \log(\sigma_{b,i}^2) - \mu_{b,i}^2 - \sigma_{b,i}^2),
    \end{aligned}
\end{equation}
where $N$ denotes the batch size, $V_0^b$ the set of nodes in top-level mesh for sample $b$, $m_b$ the corresponding active-node mask, and $V_d^b$ the set of nodes in the latent mesh for sample $b$.
\paragraph{VGAE hyper-parameters.}
VGAE hyper-parameters are summarized in Table \ref{appendix:gvae-arch-details}. 
\begin{table*}[th]
\centering
\small
\caption{VGAE hyper-parameters.} \label{appendix:gvae-arch-details}
\begin{tabular}{lr}
    \hline
    GraphPool hidden size & 32 \\
    GraphPool hidden layers & 2 \\
    GraphUnpool hidden size & 32 \\
    GraphPool hidden layers & 2 \\
    Input features size & 3 (temperature, mask, material mask) \\
    Output features size & 1 (temperature) \\
    Encoder hidden size & 12, 24, 32 \\
    Decoder hidden size & 32, 24, 12 \\
    Depth & 3\\
    $\beta$ & 100\\
    \hline
\end{tabular}
\end{table*}
\subsection{Latent-RGNN Architecture Details} \label{appendix:sec:latent-RGNN-arch}
The latent-RGNN is an extension of the RGNN (see Section \ref{sec:RGNN}) operating directly in latent space. It outputs a sequence of latent representations, which are subsequently decoded into temperature fields using the pre-trained VGAE decoder. To improve computational efficiency, latent representations are decoded in batches rather than sequentially, thereby reducing decoding overhead.
In contrast to the original RGNN formulation, which operates on full-resolution temperature fields and directly incorporates the active-node and material masks, the latent-RGNN evolves its hidden state on the latent mesh. However, the active-node and material masks remain defined on the original, high-resolution mesh. To reconcile this mismatch, we employ a lightweight graph encoder based on the graph pooling operations described in Section \ref{appendix:sec:VGAE} to project these auxiliary inputs onto the latent mesh. The resulting compressed representations are concatenated with the latent hidden state and processed by the GraphGRU (see Figure \ref{fig:framework}).
\subsection{Interlayer and Intralayer Models}
The interlayer and intralayer models are built upon the latent RGNN architecture. Since the latent graph produced by the latent RGNN must be decoded via the VGAE decoder, multiple time steps are batched before decoding to improve computational efficiency, with a marginal increase in memory usage.
Architectural hyper-parameters are summarized in Table~\ref{appendix:inter-intralayer-arch-details}.
\begin{table*}[th]
\centering
\small
\caption{Architecture details of the interlayer and intralayer models. The geometry encoder is used to encode geometrical inputs (active nodes mask, material mask) to the latent mesh. MPGNN denotes Message Passing Graph Neural Network and is used in the GraphGRU architecture.} \label{appendix:inter-intralayer-arch-details}
\begin{tabular}{lcc}
    \cline{2-3}
    & Interlayer latent RGNN & Intralayer latent RGNN \\
    \hline
    Hidden feature size & 48 & 32 \\
    Geometry encoder: GraphPool hidden size & 16 & 16 \\
    Geometry encoder: GraphPool hidden layers & 2 & 2\\
    Geometry encoder hidden size & 8, 16, 16 & 8, 16, 16\\
    Geometry encoder input features size & 2  & 2\\
    Geometry encoder output features size & 16 & 16\\
    GraphGRU MPGNNs hidden size & 48 & 32 \\
    GraphGRU MPGNNs processor blocks & 2 & 1\\
    GraphGRU MPGNNs processor depth & 2 & 2 \\
    GraphGRU MPGNNs node processor hidden layers & 2 & 2\\
    GraphGRU MPGNNs edge processor hidden layers & 2 & 2\\
    \hline
\end{tabular}
\end{table*}
\section{Baseline Decoupled-RGNN Model Implementation} \label{appendix:dec-rgnn}
For comparison, we reimplemented the RGNN model proposed by \cite{choi2025transfer}. Since the original work considers a different printing process and operates on three-dimensional data, a direct application to our setting is not possible. We therefore adapted their methodology to our dataset while remaining as faithful as possible to the original implementation.
The architecture in \cite{choi2025transfer} is based on a recurrent graph neural network that differs from ours in several key aspects. Their model processes input sequences of 50 simulation time steps, each represented as a graph encoding geometric and process-related features. Node and edge features are first embedded, and each graph is processed independently by a geometric encoder composed of stacked DeepGCN layers (see \cite{li2019deepgcns}). This constitutes the spatial modeling stage.
Subsequently, temporal modeling is performed without further use of the graph structure: the graph-encoded representations are reshaped into per-node temporal sequences and passed through a stacked GRU module, which outputs the predicted temperature sequence at each node. In this design, spatial and temporal processing are explicitly decoupled, in contrast to our approach, which jointly models spatiotemporal dependencies.
We note that the baseline model is non-autoregressive and can process arbitrary subsequences of simulation time steps without conditioning on previously predicted temperature states. For training, we initialized the model using the hyper-parameters reported in the original paper and further tuned them to optimize performance on our dataset.
A direct quantitative comparison with the results reported in \cite{choi2025transfer} is not feasible, as the datasets differ substantially and the original work does not provide sufficiently detailed performance metrics.
\section{Training Details} \label{appendix:train-details}
\paragraph{Training Setup.}
All models were trained using the SOAP optimizer~\cite{vyas2024soap} with a learning-rate schedule combining linear warm-up followed by cosine annealing \cite{loshchilov2016sgdr}. Training was performed on a single NVIDIA A100 GPU with 40GB of memory. The validation loss was monitored during training, and early stopping was applied when it ceased to improve. Recurrent models were trained using TBPTT.
The learning rate and batch size were selected to maximize validation performance while respecting GPU memory constraints. Due to the large memory footprint of the full-resolution simulation data, the batch size was strongly constrained in the full-dimensional setting. Operating in the latent space induced by the VGAE substantially alleviated memory pressure, enabling larger and more flexible batch sizes. However, latent-space models exhibited more challenging optimization behavior when trained on short temporal subsequences compared to their full-resolution RGNN counterparts. 
\paragraph{Temperature Sequence Subsampling.}
To reduce memory consumption during training and to accelerate optimization, we employed temporal subsampling strategies for the temperature sequences. For the VGAE, we retained one simulation time step out of seven, while preserving all layer-wise terminal temperature fields, which are subsequently used by the interlayer model. This sampling strategy was sufficient to maintain adequate diversity in the temperature fields. No subsampling was applied to the interlayer model, which operates on a single temperature field per printed layer. For the intralayer model, temperature sequences were subsampled at a rate of one time step out of seven, resulting in shorter subsequences that facilitated more efficient and stable training.
This subsampling scheme substantially reduces the memory footprint of the simulation data while preserving satisfactory temporal fidelity, as seven simulation steps correspond to approximately 1.75 seconds. Models trained under this regime can be applied at full temporal resolution during inference; however, this may induce a degradation in accuracy due to the resulting distribution shift. This limitation could be mitigated by training with multiple subsampling rates for the intralayer sequences, thereby improving robustness to varying temporal resolutions at inference time. Finally, we note that higher temporal resolution at inference increases computational cost, as it entails predicting a larger number of time steps. Even with a subsampling rate of one out of seven, the resulting sequences remain long, comprising approximately 1,000 to 1,700 time steps.
\paragraph{Training Hyper-parameters.}
The training hyper-parameters selected for the different model architectures are summarized in Table~\ref{appendix:training-details-hyperparams}.
\begin{table*}[h]
\centering
\small
\caption{Training hyper-parameters. The LR warmup is a linear LR scheduler with start factor = 0.1 and duration = 8.} \label{appendix:training-details-hyperparams}
\begin{tabular}{l|cccc}
    \hline
    Model & Learning rate & Batch size & LR warmup & LR scheduler \\
    \hline
    VGAE & 1e-3 & 64 & Yes & Cosine annealing ($T_\text{max}=130$, $\eta_{\text{min}} = 1e-4$) \\
    Interlayer & 2e-3 & 4 & Yes & Cosine annealing ($T_\text{max}=140$, $\eta_{\text{min}} = 1e-5$) \\
    Intralayer & 4e-3 & 64 & Yes & Cosine annealing ($T_\text{max}=180$, $\eta_{\text{min}} = 1e-5$) \\
    Decoupled-RGNN & 1e-3 & 4 & Yes & Cosine annealing ($T_\text{max}=130$, $\eta_{\text{min}} = 1e-3$)\\
    \hline
\end{tabular}
\end{table*}
The training hyper-parameters that were kept identical across all experiments are summarized in Table~\ref{appendix:training-details-fixed-hyperparams}.
\begin{table*}[h]
\centering
\small
\caption{Training hyper-parameters with a fixed value.} \label{appendix:training-details-fixed-hyperparams}
\begin{tabular}{lr}
    \hline
    TBPTT steps & 8 \\ 
    Early stopping patience & 25 \\ 
    SOAP optimizer betas & 0.95, 0.95 \\ 
    SOAP optimizer weight decay & 0.01 \\ 
    SOAP optimizer preconditioning frequency & 8 \\ \hline
\end{tabular}
\end{table*}
The TBPTT step parameter specifies the number of forward time steps processed before performing backpropagation through time. Further details on this procedure can be found in \cite{jaeger2002tutorial}. The early-stopping patience parameter determines the number of consecutive epochs with no improvement in validation loss that are tolerated before terminating training. The SOAP optimizer \citep{vyas2024soap} was selected after comparative experiments with AdamW, as it consistently yielded comparable or improved performance with negligible additional training overhead.

\section{Evaluation Metrics} \label{appendix:eval-metrics}
\subsection{Metrics Definitions}
We evaluate predictive performance using several complementary error metrics.
The Mean Absolute Error (MAE), Mean Absolute Percentage Error (MAPE), Mean Maximum Error (MME) are defined as
\begin{equation}
    \begin{aligned}
        &\text{MAE}(T^{true}, T^{pred}) = \frac{1}{N} \sum_{b=1}^N \frac{1}{|m_b|} \sum_{i \in V_0^b} m_{b,i} . | T_{b,i}^{true} - T_{b,i}^{pred} |, \\
        &\text{MAPE}(T^{true}, T^{pred}) = \frac{1}{N} \sum_{b=1}^N \frac{1}{|m_b|} \sum_{i \in V_0^b} m_{b,i} . \frac{| T_{b,i}^{true} - T_{b,i}^{pred} |}{| T_{b,i}^{true} | + \epsilon},\\
        &\text{MME}(T^{pred}) = \frac{1}{N} \sum_{b=1}^N \,\, \max_{i \in V_0^b} \, \left(  m_{b,i} . | T_{b}^{true} - T_{b}^{pred} | \right), 
    \end{aligned}
    \label{eq:metrics1}
\end{equation}
where $m_b$ denotes the active-node mask for sample $b$, $N$ is the number of samples, and $V_0^b$ is the set of mesh nodes on which the temperature fields $T^{true}$ and $T^{pred}$ are defined. \\
To assess the quality of melt-pool localization, we employ the melt-pool Intersection over Union (mpIoU), defined as
\begin{equation*}
    \begin{aligned}
        &\text{mpIoU}(T^{true}, T^{pred}) =  \frac{1}{N} \sum_{b=1}^N \frac{|\lbrace T^{true}_b > \theta_f\rbrace \cap \lbrace T^{pred}_b > \theta_f \rbrace|}{|\lbrace T^{true}_b > \theta_f\rbrace \cup \lbrace T^{pred}_b > \theta_f \rbrace|},
    \end{aligned}
\end{equation*}
where $\theta_f = 1170$\textdegree C corresponds to the melting temperature of the material considered in this study.\\
Temporal stability is quantified using the temporal MAE (t-MAE), defined as the time-resolved MAE over a temperature sequence:
\begin{equation*}
    \begin{aligned}
        &\text{t-MAE}\left( \lbrace T^{true}(t) \rbrace_{t=1..N_t}, \lbrace T^{pred}(t) \rbrace_{t=1..N_t} \right) = \Bigl\{ \sum_{b=1}^N \frac{1}{N} \,\, \text{MAE}\left(T^{true}_b(t), T^{pred}_b(t)\right) \Bigl\}_{t=1..N_t},
    \end{aligned}
\end{equation*}
where $\lbrace T^{true}(t) \rbrace_{t=1..N_t}$ and $\lbrace T^{pred}(t) \rbrace_{t=1..N_t}$ are true and predicted temperature fields sequence respectively and $N_t$ denotes the sequence length.\\
Finally, we evaluate errors in spatial and temporal temperature gradients through the Mean Absolute Error of spatial gradients ($\text{MAE}_{\nabla_xy}$) and temporal gradients ($\text{MAE}_{\nabla_t}$), defined as
\begin{equation*}
    \begin{aligned}
        &\text{MAE}_{\nabla_{xy}}(T^{true}, T^{pred}) = \frac{1}{N} \sum_{b=1}^N \frac{1}{|m_b|} \sum_{i \in V_0^b} m_{b,i} . \frac{1}{2} \left(| \partial_x T_{b,i}^{true} - \partial_x T_{b,i}^{pred} | + | \partial_y T_{b,i}^{true} - \partial_y T_{b,i}^{pred} |\right), \\
        &\text{MAE}_{\nabla_t}(T^{true}, T^{pred}) = \frac{1}{N} \sum_{b=1}^N \frac{1}{|m_b|} \sum_{i \in V_0^b} m_{b,i} . | \partial_t T_{b,i}^{true} - \partial_t T_{b,i}^{pred} |,
    \end{aligned}
\end{equation*}
where $\partial_x$ and $\partial_y$ denote spatial derivatives along the $x$ and $y$ directions, computed on quadrilateral mesh elements and then projected onto nodes, and $\partial_t$ denotes the temporal derivative, computed using finite differences.
\paragraph{Inference Time.}
The inference times reported in Table~\ref{tab:models-perfs2} correspond to per-time-step inference costs. These values were obtained by evaluating each model sequentially on the 40 test simulations, measuring the total inference time per simulation, and normalizing by the number of predicted time steps. To ensure reliable measurements, an initial warm-up pass was performed without timing, followed by three full evaluations on the test set. The reported inference times are computed as the average over these runs.
\paragraph{Memory Pressure.}
In our setting, memory footprint is difficult to compare quantitatively across models, as it varies substantially with batch size, architectural choices, and other hyper-parameters. We therefore report a qualitative assessment in Table~\ref{tab:models-perfs2}, based on the practical difficulty of fitting each model within the 40GB GPU memory limit while maintaining stable training and satisfactory convergence. Although this measure does not precisely reflect absolute memory consumption, it provides useful insight into the impact of memory constraints when selecting model architectures and tuning their hyper-parameters.
\subsection{Relevance of Physical Evaluation Metrics} \label{appendix:metrics-phy-relevance}
The physically motivated metrics (mpIoU, $\text{MAE}_{\nabla_{xy}}$, and $\text{MAE}_{\nabla_t}$) are particularly relevant, as melt-pool extent and temperature gradients in space and time directly influence microstructure evolution and residual stresses in additively manufactured parts \cite{bian2025review}. These quantities are therefore critical for assessing part quality and expected service life: microstructure governs material properties, while residual stresses affect fatigue resistance. Accurate prediction under these metrics supports process optimization toward reduced residual stresses and enhanced microstructural homogeneity. \cite{ali2018residual}.
\section{Experiments \& Results Supplementary Material} \label{appendix:results-details}
\subsection{On the Early-Time MAE Peak in the Temporal MAE Analysis} \label{appendix:tmae-peak}
\begin{figure}[!htb]
    \centering
    \includegraphics[width=0.45\linewidth]{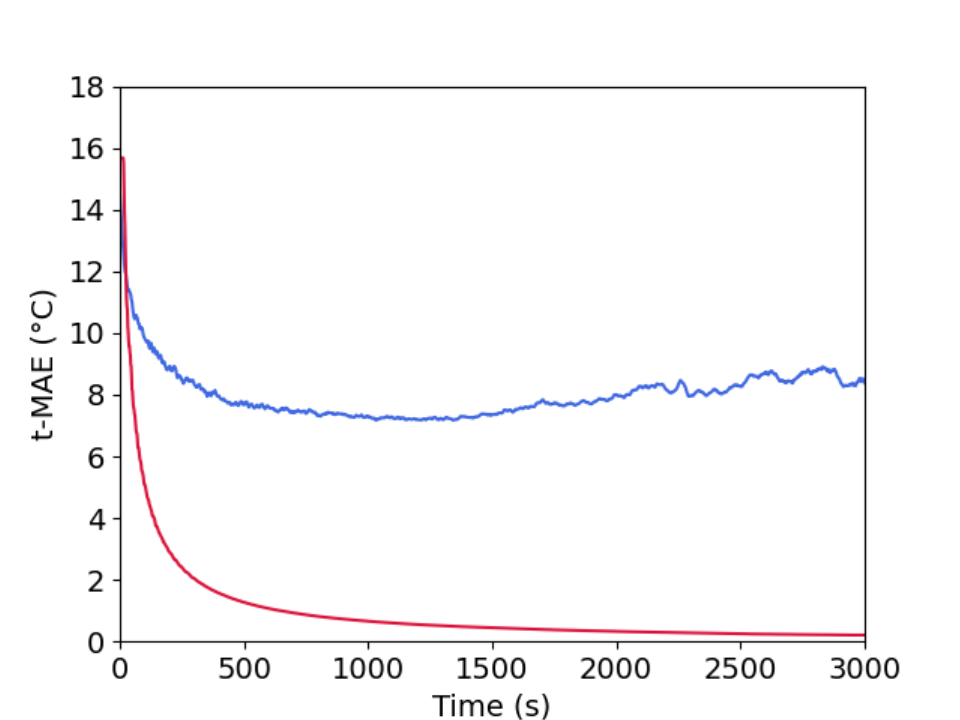}
    \caption{The blue curve is the temporal MAE (t-MAE) on the test set for the VGAE with latent dim. 4. The red curve is the function $f(t) = \text{t-MAE}(0) / \text{domain size(t)}$.} 
    \label{fig:tmae-4-inv-domain-size}
\end{figure}

Figure~\ref{fig:tmae-4-inv-domain-size} reports the temporal MAE (t-MAE) of the VGAE with latent dimensionality 4 (blue curve), together with the reference function $f(t) = \text{t-MAE}(0) / \text{domain size(t)}$ (red curve), where $\text{t-MAE}(0)$ denotes the MAE of the VGAE at $t=0$ and $\text{domain size(t)}$ corresponds to the mean number of active mesh nodes over the test set at time $t$. The function $f(t)$ represents the expected decay of the MAE under the idealized assumption that the prediction error remains constant in magnitude but is spatially concentrated at a single location, so that its average decreases solely due to the growth of the active domain.
This comparison partially explains the observed shape of the VGAE t-MAE curve. In practice, temperature prediction errors are spatially localized around the laser path, where temperature gradients are highest, particularly at early time steps when thermal diffusion has had limited effect. As the active domain expands over time, these localized errors are progressively diluted, leading to a decreasing average MAE.
A similar behavior is observed across all models, including the baseline and ablated variants. Notably, even for the interlayer model (where predictions correspond to cooling phases with the laser turned off) an initial MAE peak persists. This effect can be attributed to residual error localization in specific regions of the part, such as the upper layers of the metallic domain.
Overall, the early-time MAE peak should be interpreted as a consequence of the temporal evolution of the active domain size, rather than as evidence of systematically degraded predictive performance at early stages of the printing process.

\subsection{VGAE Model Performance} \label{appendix:sec:VGAE-perfs}
\begin{figure}[!htb]
    \centering
    \includegraphics[width=0.45\linewidth]{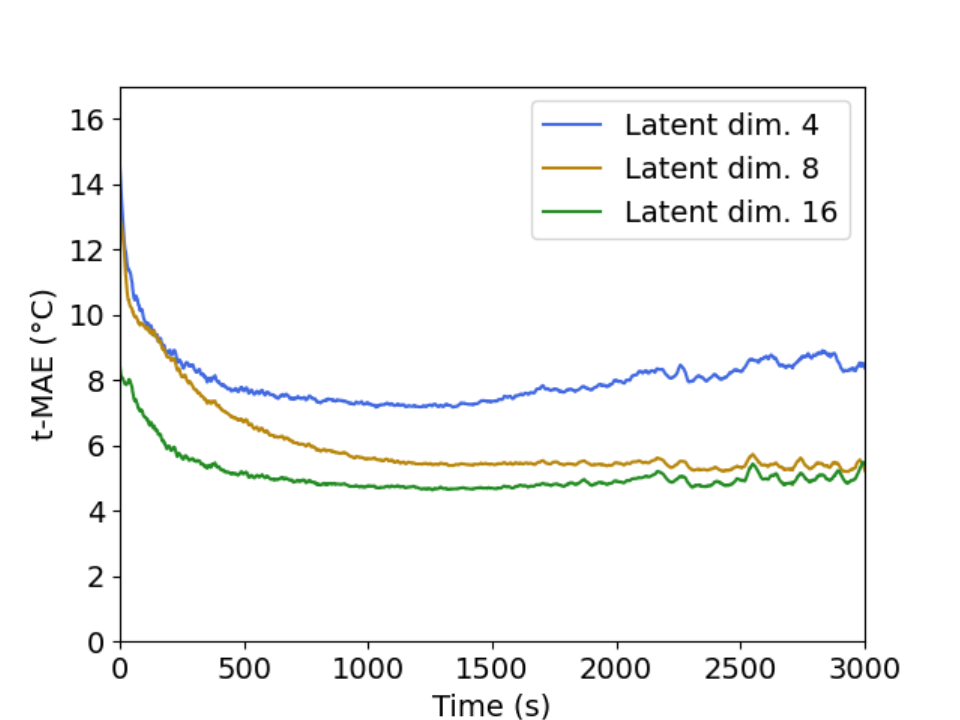}
    \caption{Temporal MAE of the VGAE on the test set for different latent dimensionalities. The initial peak in error is attributable to the concentration of reconstruction errors in the vicinity of the laser beam combined with the small domain size at early time steps, as discussed in Appendix \ref{appendix:tmae-peak}.}
    \label{fig:tmae-vgae}
\end{figure}

Tables~\ref{tab:VGAE-perfs} and~\ref{tab:VGAE-comp-perfs} report, respectively, the reconstruction performance metrics and compression characteristics of the VGAE on the test set for varying latent dimensionalities. As the latent dimension increases, reconstruction accuracy improves, while the memory reduction factor correspondingly decreases. The memory reduction factor is computed as the ratio between the total storage required for a full simulation and its compressed latent counterpart.
Although increasing the latent dimensionality increases the per-node feature size, this is partially compensated by the substantial reduction in mesh resolution achieved by the encoder. In the present configuration, a VGAE depth of three results in an isotropic reduction of the mesh resolution by a factor of two, corresponding to an overall reduction factor of 64 in the number of nodes.
The observed MME may appear relatively large; however, even minor spatial deviations in melt pool localization can induce large pointwise errors due to the extremely steep temperature gradients in the vicinity of the melt pool.
\begin{table*}[t]
\centering
\small
\caption{VGAE performance on the test set for varying latent space dimensionalities.} \label{tab:VGAE-perfs}
\begin{tabular}{c|ccccccc}  
    \hline
    Latent dim. & MAE (°C) & MAPE (\%) & MME (°C) & mpIoU & $\text{MAE}_{\nabla_{xy}}$ (°C/mm) & $\text{MAE}_{\nabla_{t}}$ (°C/s) & Parameters\\
    \hline
    4 & 7.05 $\pm$ 1.68 & 2.44 $\pm$ 0.51 & 118.6 $\pm$ 22.5 & 0.86 $\pm$ 0.02 & 3.01 $\pm$ 0.30 & 3.68 $\pm$ 0.84 & 19.0k \\
    8 & 5.74 $\pm$ 0.93 & 2.08 $\pm$ 0.31 & 98.3 $\pm$ 12.7  & 0.88 $\pm$ 0.003 & 2.62 $\pm$ 0.28 & 2.86 $\pm$ 0.32 & 19.8k \\
    16 & \textbf{5.08 $\pm$ 0.03} & \textbf{1.82 $\pm$ 0.10} & \textbf{94.5 $\pm$ 2.5} & \textbf{0.89 $\pm$ 0.003} & \textbf{2.35 $\pm$ 0.005} & \textbf{2.59 $\pm$ 0.05} & 21.5k \\
    \hline
\end{tabular}
\end{table*}
\begin{table}[ht]
\centering
\small
\caption{VGAE compression performance on the test set for varying latent space dimensionalities. “Sim. data” refers to the combined storage of the simulation mesh description and the corresponding sequence of temperature fields. \label{tab:VGAE-comp-perfs}}
\begin{tabular}{c|cc} 
    \hline
    Latent dim. & Sim. data footprint & Mem. reduction factor \\
    \hline
    4 &  \textbf{92 MB}  & \textbf{0.06}  \\
    8 &  185 MB & 0.12 \\
    16 & 369 MB & 0.23 \\
    No compress. & 795 MB &  \\
    \hline
\end{tabular}
\end{table}
Although the VGAE does not explicitly model temporal dynamics, its reconstruction error can still be analyzed as a function of time. Figure~\ref{fig:tmae-vgae} reports the t-MAE across simulation time steps. An error peak is observed during the early stages of the printing process; however, this does not indicate a systematic degradation of model performance at early steps. This behavior arises from the definition of the MAE, which is computed exclusively over active nodes (see Eq.~\ref{eq:metrics1} in Appendix \ref{appendix:eval-metrics}). More details are given in Appendix \ref{appendix:tmae-peak}. 

\subsection{Temperature predictions} \label{appendix:temperature-figd}
\begin{figure}[!htb]
    \centering
    \begin{tabular}{|C{0.5cm}|C{0.5cm}|C{12cm}|}
      \hline
      \centering
      & \centering\rotatebox{90}{True temperature (°C)} & \adjustbox{center, padding=2pt}{\includegraphics[width=0.7\textwidth]{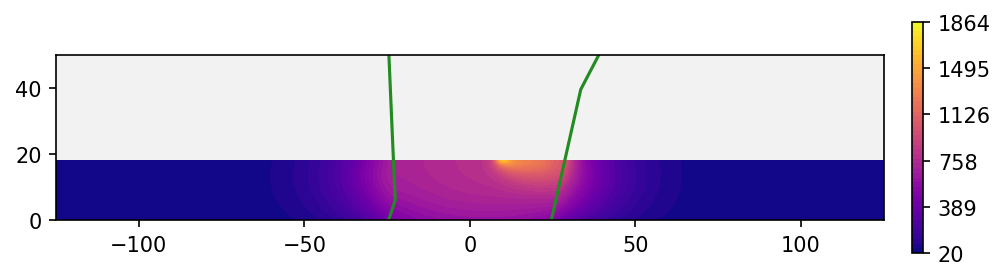}} \\
      \hline
      \multirow[c]{2}{*}{\centering\rotatebox{90}{LM-RGNN model}} & \centering\rotatebox{90}{Prediction (°C)} & \adjustbox{center, padding=2pt}{\includegraphics[width=0.7\textwidth]{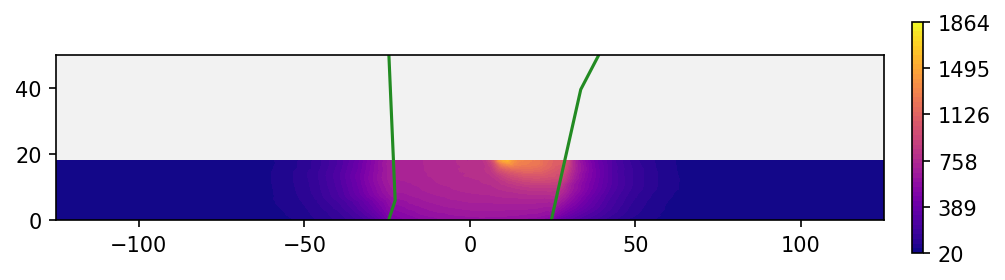}}  \\
      \cline{2-3}
      & \centering\rotatebox{90}{MAE (°C)} & \adjustbox{center, padding=2pt}{\includegraphics[width=0.7\textwidth]{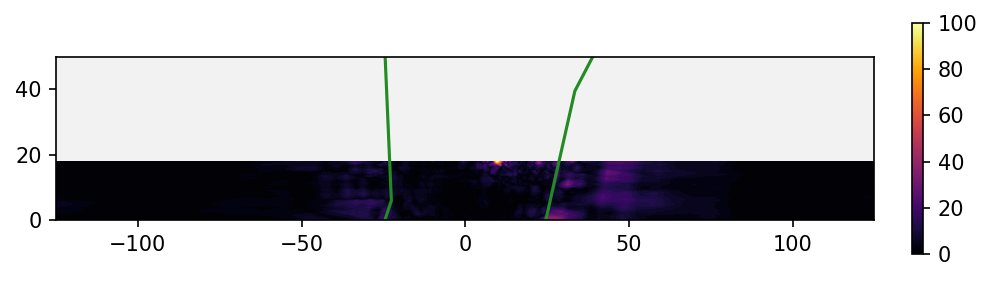}} \\
      \hline
      \multirow[c]{2}{*}{\centering\rotatebox{90}{Decoupled-RGNN model}} & \centering\rotatebox{90}{Prediction (°C)} & \adjustbox{center, padding=2pt}{\includegraphics[width=0.7\textwidth]{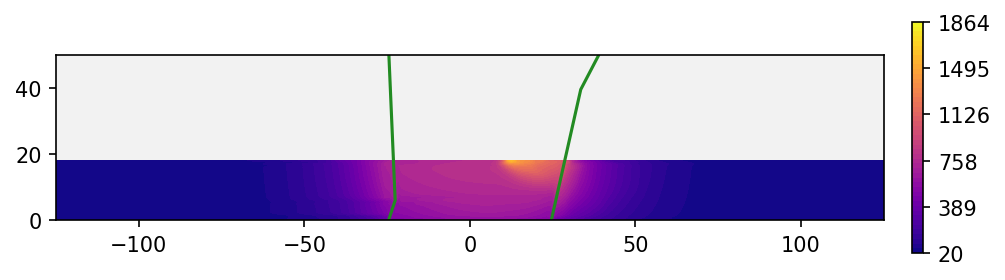}} \\
      \cline{2-3}
      & \centering\rotatebox{90}{MAE (°C)} & \adjustbox{center, padding=2pt}{\includegraphics[width=0.7\textwidth]{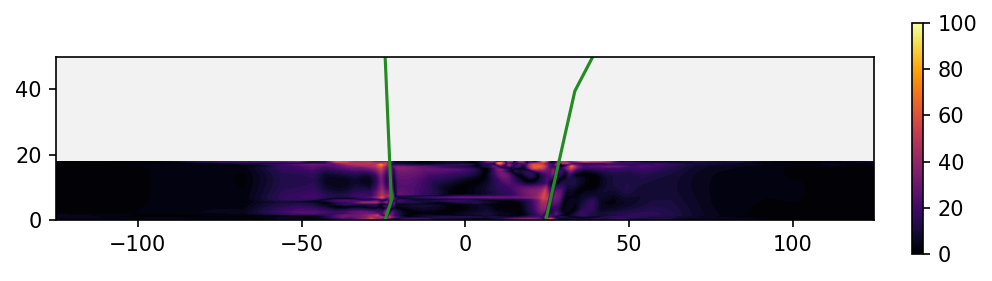}} \\
      \hline
    \end{tabular}
    \caption{Comparison of ground-truth temperature fields with predictions from the LM-RGNN and the Decoupled-RGNN for one of the test geometries at $t=700$s. The corresponding temperature MAE is reported for each model. For visualization, the MAE color scale is clipped at 100°C, as the Decoupled-RGNN can produce substantially higher maximum errors compared to the LM-RGNN. The green lines indicate the part boundaries, enclosing the solid metallic region.}
    \label{fig:temperature-array}
\end{figure}

\begin{figure}[!htb]
    \centering
    \begin{tabular}{|C{0.5cm}|C{0.5cm}|C{12cm}|}
      \hline
      \centering
      & \centering\rotatebox{90}{True temperature (°C)} & \adjustbox{center, padding=2pt}{\includegraphics[width=0.7\textwidth]{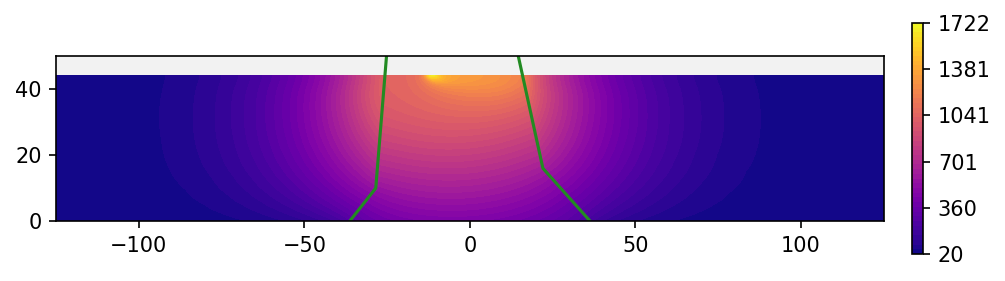}} \\
      \hline
      \multirow[c]{2}{*}{\centering\rotatebox{90}{LM-RGNN model}} & \centering\rotatebox{90}{Prediction (°C)} & \adjustbox{center, padding=2pt}{\includegraphics[width=0.7\textwidth]{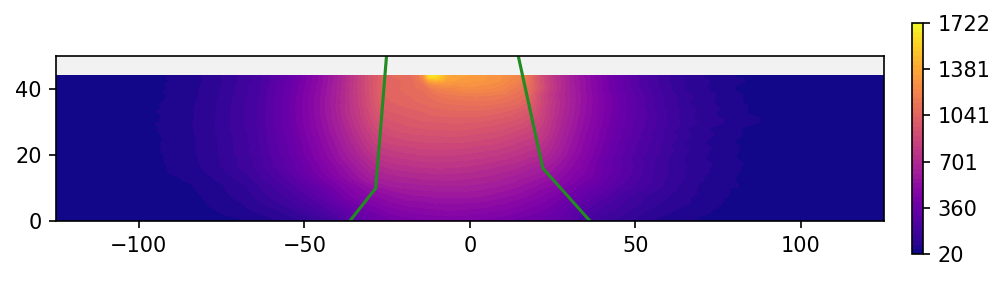}}  \\
      \cline{2-3}
      & \centering\rotatebox{90}{MAE (°C)} & \adjustbox{center, padding=2pt}{\includegraphics[width=0.7\textwidth]{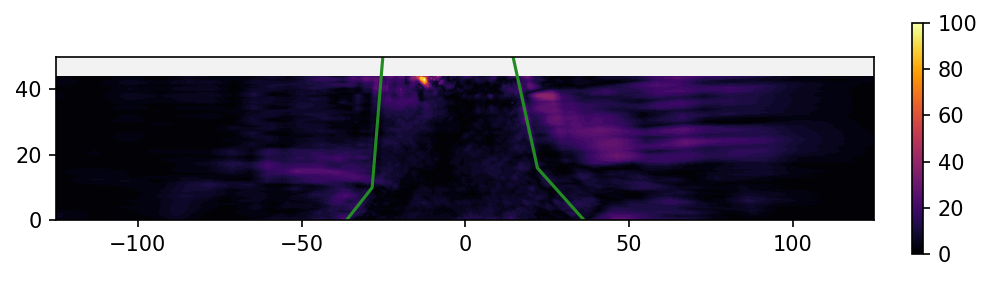}} \\
      \hline
      \multirow[c]{2}{*}{\centering\rotatebox{90}{Decoupled-RGNN model}} & \centering\rotatebox{90}{Prediction (°C)} & \adjustbox{center, padding=2pt}{\includegraphics[width=0.7\textwidth]{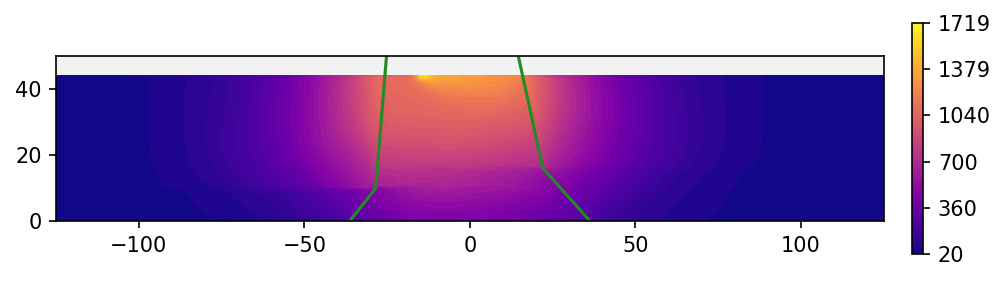}} \\
      \cline{2-3}
      & \centering\rotatebox{90}{MAE (°C)} & \adjustbox{center, padding=2pt}{\includegraphics[width=0.7\textwidth]{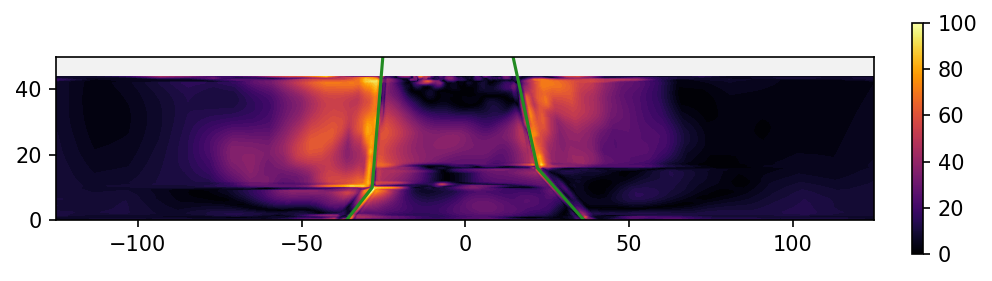}} \\
      \hline
    \end{tabular}
    \caption{Comparison of ground-truth temperature fields with predictions from the LM-RGNN and the Decoupled-RGNN for one of the test geometries at $t=1800$s. The corresponding temperature MAE is reported for each model. For visualization, the MAE color scale is clipped at 100°C, as the Decoupled-RGNN can produce substantially higher maximum errors compared to the LM-RGNN. The green lines indicate the part boundaries, enclosing the solid metallic region.}
    \label{fig:temperature-array2}
\end{figure}

Figures~\ref{fig:temperature-array} and~\ref{fig:temperature-array2} present temperature field predictions obtained with the LM-RGNN and the Decoupled-RGNN models, respectively.

%% file: paper.bib
@inproceedings{
    valencia2025learning,
    title={Learning Distributions of Complex Fluid Simulations with Diffusion Graph Networks},
    author={Mario Lino Valencia and Tobias Pfaff and Nils Thuerey},
    booktitle={The Thirteenth International Conference on Learning Representations},
    year={2025},
    url={https://openreview.net/forum?id=uKZdlihDDn}
}

@article{kipf2016variational,
  title={Variational graph auto-encoders},
  author={Kipf, Thomas N and Welling, Max},
  journal={arXiv preprint arXiv:1611.07308},
  year={2016}
}

@article{brody2021attentive,
  title={How attentive are graph attention networks?},
  author={Brody, Shaked and Alon, Uri and Yahav, Eran},
  journal={arXiv preprint arXiv:2105.14491},
  year={2021}
}

@inproceedings{higgins2017beta,
  title={beta-vae: Learning basic visual concepts with a constrained variational framework},
  author={Higgins, Irina and Matthey, Loic and Pal, Arka and Burgess, Christopher and Glorot, Xavier and Botvinick, Matthew and Mohamed, Shakir and Lerchner, Alexander},
  booktitle={International conference on learning representations},
  year={2017}
}

@article{vyas2024soap,
  title={Soap: Improving and stabilizing shampoo using adam},
  author={Vyas, Nikhil and Morwani, Depen and Zhao, Rosie and Kwun, Mujin and Shapira, Itai and Brandfonbrener, David and Janson, Lucas and Kakade, Sham},
  journal={arXiv preprint arXiv:2409.11321},
  year={2024}
}

@article{choi2025transfer,
  title={Transfer learning enabled geometry, process, and material agnostic {RGNN} for temperature prediction in directed energy deposition},
  author={Choi, Jin Young and Estalaki, Sina Malakpour and Quispe, Daniel and Zha, Rujing and Rolark, Rowan and Mozaffar, Mojtaba and Cao, Jian},
  journal={Additive Manufacturing},
  pages={104876},
  year={2025},
  publisher={Elsevier}
}

@inproceedings{li2019deepgcns,
  title={Deep{GCN}s: {C}an {GCN}s go as deep as {CNN}s?},
  author={Li, Guohao and Muller, Matthias and Thabet, Ali and Ghanem, Bernard},
  booktitle={Proceedings of the IEEE/CVF international conference on computer vision},
  pages={9267--9276},
  year={2019}
}

@article{lippe2023pde,
  title={Pde-refiner: Achieving accurate long rollouts with neural pde solvers},
  author={Lippe, Phillip and Veeling, Bas and Perdikaris, Paris and Turner, Richard and Brandstetter, Johannes},
  journal={Advances in Neural Information Processing Systems},
  volume={36},
  pages={67398--67433},
  year={2023}
}

@article{loshchilov2016sgdr,
  title={Sgdr: Stochastic gradient descent with warm restarts},
  author={Loshchilov, Ilya and Hutter, Frank},
  journal={arXiv preprint arXiv:1608.03983},
  year={2016}
}

@book{jaeger2002tutorial,
  title={Tutorial on training recurrent neural networks, covering {BPPT}, {RTRL}, {EKF} and the echo state network approach},
  author={Jaeger, Herbert},
  volume={5},
  year={2002},
  publisher={GMD-Forschungszentrum Informationstechnik Bonn}
}

@article{price2025probabilistic,
  title={Probabilistic weather forecasting with machine learning},
  author={Price, Ilan and Sanchez-Gonzalez, Alvaro and Alet, Ferran and Andersson, Tom R and El-Kadi, Andrew and Masters, Dominic and Ewalds, Timo and Stott, Jacklynn and Mohamed, Shakir and Battaglia, Peter and others},
  journal={Nature},
  volume={637},
  number={8044},
  pages={84--90},
  year={2025},
  publisher={Nature Publishing Group UK London}
}

@article{tian2025physics,
  title={Physics-informed machine learning-based real-time long-horizon temperature fields prediction in metallic additive manufacturing},
  author={Tian, Mingxuan and Mu, Haochen and Liu, Tao and Li, Mengjiao and Ding, Donghong and Zhao, Jianping},
  journal={Communications Engineering},
  volume={4},
  number={1},
  pages={168},
  year={2025},
  publisher={Nature Publishing Group UK London}
}

@article{gao2024towards,
  title={Towards spatio-temporal prediction of cavitating fluid flow with graph neural networks},
  author={Gao, Rui and Heydari, Shayan and Jaiman, Rajeev K},
  journal={International Journal of Multiphase Flow},
  volume={177},
  pages={104858},
  year={2024},
  publisher={Elsevier}
}

@article{yue2024tgn,
  title={{TGN}: A Temporal Graph Network for Physics Prediction},
  author={Yue, Miaocong and Liu, Huayong and Chang, Xinghua and Zhang, Laiping and Li, Tianyu},
  journal={Applied Sciences},
  volume={14},
  number={2},
  pages={863},
  year={2024},
  publisher={MDPI}
}

@article{zhu2021machine,
  title={Machine learning for metal additive manufacturing: predicting temperature and melt pool fluid dynamics using physics-informed neural networks},
  author={Zhu, Qiming and Liu, Zeliang and Yan, Jinhui},
  journal={Computational Mechanics},
  year={2021},
  volume={67},
  number={2},
  pages={619--635},
  publisher={Springer}
}

@article{janny2023eagle,
  title={Eagle: Large-scale learning of turbulent fluid dynamics with mesh transformers},
  author={Janny, Steeven and Beneteau, Aur{\'e}lien and Nadri, Madiha and Digne, Julie and Thome, Nicolas and Wolf, Christian},
  journal={arXiv preprint arXiv:2302.10803},
  year={2023}
}

@inproceedings{karkaria2025asno,
  title={{ASNO}: An Interpretable Attention-Based Spatio-Temporal Neural Operator for Robust Scientific Machine Learning}, year={2025},
  author={Karkaria, Vispi Nevile and Lee, Doksoo and Chen, Yi-Ping and Yu, Yue and Chen, Wei},
  booktitle={ICML 2025 Workshop on Reliable and Responsible Foundation Models}
}

@article{liao2023hybrid,
  title={Hybrid thermal modeling of additive manufacturing processes using physics-informed neural networks for temperature prediction and parameter identification},
  author={Liao, Shuheng and Xue, Tianju and Jeong, Jihoon and Webster, Samantha and Ehmann, Kornel and Cao, Jian},
  journal={Computational Mechanics},
  volume={72},
  number={3},
  pages={499--512},
  year={2023},
  publisher={Springer}
}

@article{verma2024climode,
  title={Climode: Climate and weather forecasting with physics-informed neural odes},
  author={Verma, Yogesh and Heinonen, Markus and Garg, Vikas},
  journal={arXiv preprint arXiv:2404.10024},
  year={2024}
}

@article{zhao2019t,
  title={{T-GCN}: A temporal graph convolutional network for traffic prediction},
  author={Zhao, Ling and Song, Yujiao and Zhang, Chao and Liu, Yu and Wang, Pu and Lin, Tao and Deng, Min and Li, Haifeng},
  journal={IEEE transactions on intelligent transportation systems},
  volume={21},
  number={9},
  pages={3848--3858},
  year={2019},
  publisher={IEEE}
}

@article{narasimharaju2022comprehensive,
  title={A comprehensive review on laser powder bed fusion of steels: Processing, microstructure, defects and control methods, mechanical properties, current challenges and future trends},
  author={Narasimharaju, Shubhavardhan Ramadurga and Zeng, Wenhan and See, Tian Long and Zhu, Zicheng and Scott, Paul and Jiang, Xiangqian and Lou, Shan},
  journal={Journal of Manufacturing Processes},
  volume={75},
  pages={375--414},
  year={2022},
  publisher={Elsevier}
}

@article{chiumenti2017numerical,
  title={Numerical modelling and experimental validation in Selective Laser Melting},
  author={Chiumenti, Michele and Neiva, Eric and Salsi, Emilio and Cervera, Miguel and Badia, Santiago and Moya, Joan and Chen, Zhuoer and Lee, Caroline and Davies, Christopher},
  journal={Additive Manufacturing},
  volume={18},
  pages={171--185},
  year={2017},
  publisher={Elsevier}
}

@article{burkhardt2022thermo,
  title={Thermo-mechanical simulations of powder bed fusion processes: accuracy and efficiency},
  author={Burkhardt, Christian and Steinmann, Paul and Mergheim, Julia},
  journal={Advanced Modeling and Simulation in Engineering Sciences},
  volume={9},
  number={1},
  pages={18},
  year={2022},
  publisher={Springer}
}

@article{bian2025review,
  title={A Review of the Evolution of Residual Stresses in Additive Manufacturing During Selective Laser Melting Technology},
  author={Bian, Peiying and Jammal, Ali and Xu, Kewei and Ye, Fangxia and Zhao, Nan and Song, Yun},
  journal={Materials},
  volume={18},
  number={8},
  pages={1707},
  year={2025},
  publisher={MDPI}
}

@article{abolhasani2019analysis,
  title={Analysis of melt-pool behaviors during selective laser melting of {AISI} 304 stainless-steel composites},
  author={Abolhasani, Daniyal and Seyedkashi, SM Hossein and Kang, Namhyun and Kim, Yang Jin and Woo, Young Yun and Moon, Young Hoon},
  journal={Metals},
  volume={9},
  number={8},
  pages={876},
  year={2019},
  publisher={MDPI}
}

@article{ali2018residual,
  title={Residual stress development in selective laser-melted {T}i6{A}l4{V}: a parametric thermal modelling approach},
  author={Ali, Haider and Ghadbeigi, Hassan and Mumtaz, Kamran},
  journal={The International Journal of Advanced Manufacturing Technology},
  volume={97},
  number={5},
  pages={2621--2633},
  year={2018},
  publisher={Springer}
}

@inproceedings{seo2018structured,
  title={Structured sequence modeling with graph convolutional recurrent networks},
  author={Seo, Youngjoo and Defferrard, Micha{\"e}l and Vandergheynst, Pierre and Bresson, Xavier},
  booktitle={International conference on neural information processing},
  pages={362--373},
  year={2018},
  organization={Springer}
}

@article{candreva2022current,
  title={Current and future applications of computational fluid dynamics in coronary artery disease},
  author={Candreva, Alessandro and De Nisco, Giuseppe and Rizzini, Maurizio Lodi and D’Ascenzo, Fabrizio and De Ferrari, Gaetano Maria and Gallo, Diego and Morbiducci, Umberto and Chiastra, Claudio},
  journal={Reviews in Cardiovascular Medicine},
  volume={23},
  number={11},
  pages={377},
  year={2022}
}

@article{saka2007latinized,
  title={Latinized, improved {LHS}, and {CVT} point sets in hypercubes},
  author={Saka, Yuki and Gunzburger, Max and Burkardt, John},
  journal={International Journal of Numerical Analysis and Modeling},
  volume={4},
  number={3-4},
  pages={729--743},
  year={2007}
}

@article{psiuk2024methodology,
  title={Methodology of generation of {CFD} meshes and 4{D} shape reconstruction of coronary arteries from patient-specific dynamic {CT}},
  author={Psiuk-Maksymowicz, Krzysztof and Borys, Damian and Melka, Bartlomiej and Gracka, Maria and Adamczyk, Wojciech P and Rojczyk, Marek and Wasilewski, Jaroslaw and G{\l}owacki, Jan and Kruk, Mariusz and Nowak, Marcin and others},
  journal={Scientific Reports},
  volume={14},
  number={1},
  pages={2201},
  year={2024},
  publisher={Nature Publishing Group UK London}
}

@inproceedings{ni2023voxel2hemodynamics,
  title={Voxel2hemodynamics: An end-to-end deep learning method for predicting coronary artery hemodynamics},
  author={Ni, Ziyu and Wei, Linda and Xu, Lijian and Xia, Qing and Li, Hongsheng and Zhang, Shaoting and Metaxas, Dimitris},
  booktitle={International Workshop on Statistical Atlases and Computational Models of the Heart},
  pages={15--24},
  year={2023},
  organization={Springer}
}

@article{nguyen2023climax,
  title={Climax: A foundation model for weather and climate},
  author={Nguyen, Tung and Brandstetter, Johannes and Kapoor, Ashish and Gupta, Jayesh K and Grover, Aditya},
  journal={arXiv preprint arXiv:2301.10343},
  year={2023}
}

@article{de2023machine,
  title={Machine learning for numerical weather and climate modelling: a review},
  author={de Burgh-Day, Catherine O and Leeuwenburg, Tennessee},
  journal={Geoscientific Model Development},
  volume={16},
  number={22},
  pages={6433--6477},
  year={2023},
  publisher={Copernicus Publications G{\"o}ttingen, Germany}
}

@article{suk2024mesh,
  title={Mesh neural networks for {SE} (3)-equivariant hemodynamics estimation on the artery wall},
  author={Suk, Julian and de Haan, Pim and Lippe, Phillip and Brune, Christoph and Wolterink, Jelmer M},
  journal={Computers in biology and medicine},
  volume={173},
  pages={108328},
  year={2024},
  publisher={Elsevier}
}

@article{suk2025deep,
  title={Deep vectorised operators for pulsatile hemodynamics estimation in coronary arteries from a steady-state prior},
  author={Suk, Julian and Nannini, Guido and Rygiel, Patryk and Brune, Christoph and Pontone, Gianluca and Redaelli, Alberto and Wolterink, Jelmer M},
  journal={Computer methods and programs in biomedicine},
  pages={108958},
  year={2025},
  publisher={Elsevier}
}

@article{tanade2024establishing,
  title={Establishing the longitudinal hemodynamic mapping framework for wearable-driven coronary digital twins},
  author={Tanade, Cyrus and Khan, Nusrat Sadia and Rakestraw, Emily and Ladd, William D and Draeger, Erik W and Randles, Amanda},
  journal={NPJ Digital Medicine},
  volume={7},
  number={1},
  pages={236},
  year={2024},
  publisher={Nature Publishing Group UK London}
}

@article{boutsianis2004computational,
  title={Computational simulation of intracoronary flow based on real coronary geometry},
  author={Boutsianis, Evangelos and Dave, Hitendu and Frauenfelder, Thomas and Poulikakos, Dimos and Wildermuth, Simon and Turina, Marko and Ventikos, Yiannis and Zund, Gregor},
  journal={European journal of Cardio-thoracic Surgery},
  volume={26},
  number={2},
  pages={248--256},
  year={2004},
  publisher={Elsevier Science BV}
}

@article{lam2023learning,
  title={Learning skillful medium-range global weather forecasting},
  author={Lam, Remi and Sanchez-Gonzalez, Alvaro and Willson, Matthew and Wirnsberger, Peter and Fortunato, Meire and Alet, Ferran and Ravuri, Suman and Ewalds, Timo and Eaton-Rosen, Zach and Hu, Weihua and others},
  journal={Science},
  volume={382},
  number={6677},
  pages={1416--1421},
  year={2023},
  publisher={American Association for the Advancement of Science}
}

@inproceedings{holmes2011unsteady,
  title={Unsteady vs. steady turbomachinery flow analysis: exploiting large-scale computations to deepen our understanding of turbomachinery flows},
  author={Holmes, D Graham and Moore, Branden J and Connell, Stuart D},
  booktitle={SciDAC Conference},
  year={2011}
}
